\documentclass{article}

\PassOptionsToPackage{numbers}{natbib} 
\usepackage[eandd, preprint]{neurips_2026}
\usepackage[utf8]{inputenc}   
\usepackage{rotating}
\usepackage[T1]{fontenc}      
\usepackage{hyperref}         
\usepackage{url}              
\usepackage{booktabs}         
\usepackage{amsfonts}         
\usepackage{nicefrac}         
\usepackage{microtype}        
\usepackage{xcolor}           

\usepackage{amsmath}          
\usepackage{graphicx}
\usepackage{subcaption}
\usepackage{multirow}
\usepackage{array}
\usepackage{longtable}
\usepackage{makecell}
\usepackage{float}            
\usepackage{tikz}
\usepackage{etoolbox}  
\usepackage{lineno}
\newif\ifWasLineNumbers
\AtBeginEnvironment{sidewaystable}{%
  \ifLineNumbers\WasLineNumberstrue\else\WasLineNumbersfalse\fi
  \nolinenumbers
}
\AfterEndEnvironment{sidewaystable}{%
  \ifWasLineNumbers\linenumbers\fi
}
\usetikzlibrary{positioning}

\newcommand{\metrixDoubleRelativefull}{Court Coverage Efficacy Index}
\newcommand{\metrixDoubleRelativeshort}{CCEI}
\newcommand{\metrixDefcoordinationfull}{Threat Load Asymmetry Index}
\newcommand{\metrixDefcoordinationshort}{TLAI}
\newcommand{\metrixDForcastingFs}{F1-score}
\newcommand{\metrixDForcastingAcc}{Accuracy}
\newcommand{\metrixDForcastingfullF}{Final Displacement Error}
\newcommand{\metrixDForcastingshortF}{FDE}
\newcommand{\metricDForcastingfullM}{Mean Squared Error}
\newcommand{\metricDForcastingshortM}{MSE}

\newcommand{\attackform}{offensive formation}
\newcommand{\defendform}{defensive formation}

\newcommand{\DatasetName}{DShuttleNet}

\title{\DatasetName: A Badminton Doubles Dataset for Evaluating Team Coordination}

\author{%
  Guan-Yi Jhang \\
  National Yang Ming Chiao Tung University \\
  \texttt{jhangguanyi.c@nycu.edu.tw}
  \And
  Chiencheng Chiang \\
  National Yang Ming Chiao Tung University \\
  \texttt{chienchiang.cs14@nycu.edu.tw}
  \And
  Wei-Kai Chen \\
  National Yang Ming Chiao Tung University \\
  \texttt{hyes96361@gmail.com}
  \And
  Hui-Ching Yu \\
  National Yang Ming Chiao Tung University \\
  \texttt{lillian.hc.yu.cs11@nycu.edu.tw}
  \And
  Hsu-Chao Lai \\
  Chang Gung University \\
  \texttt{hsuchaolai@cgu.edu.tw}
  \And
  Hwai-Jung Hsu \\
  Feng Chia University \\
  \texttt{hjhsu@mail.fcu.edu.tw}
  \And
  Wang Chih-Chuan \\
  National Yang Ming Chiao Tung University \\
  \texttt{wangcc@nycu.edu.tw}
  \And
  Jhy-Cheng Huang \\
  National Yang Ming Chiao Tung University \\
  \texttt{jchuang@nycu.edu.tw}
  \And
  Jiun-Long Huang \\
  National Yang Ming Chiao Tung University \\
  \texttt{jlhuang@nycu.edu.tw}
}

\begin{document}

\maketitle

\begin{abstract}
Most existing sports datasets are individual-centric or ball-centric, focusing on individual player actions while ignoring interactions among teammates. In badminton doubles, tactical success is governed by rapid team formation and coordination, yet standard resources for modeling these player interactions are scarce. To bridge this gap, we present \DatasetName, a comprehensive dataset for badminton doubles curated from elite BWF World Tour tournaments. Beyond trajectory data, \DatasetName~uniquely incorporates expert-labeled "Loss Reasons," transforming the dataset into a diagnostic tool for causal tactical evaluation. We formally define two geometry-driven representations, Teammate-Relative Angle (TRA) and Vertical Shot Angle (VSA), to characterize coordination and shot aggressiveness. To quantify interpersonal synergy, we introduce novel metrics, \metrixDoubleRelativefull~(\metrixDoubleRelativeshort) and \metrixDefcoordinationfull~(\metrixDefcoordinationshort), which evaluate the often-overlooked non-hitting partner's role and defensive coordination. Experimental results demonstrate that conventional models suffer from underfitting when failing to incorporate these coordination-centric features, highlighting the critical importance and potential impact of releasing \DatasetName. Dataset is available at: https://doi.org/10.34740/kaggle/dsv/16097938.
\end{abstract}

\vspace{-0.5em}
\section{Introduction}
\label{sec:introduction}
\vspace{-0.5em}
In the field of team sports science, the interaction between players is the fundamental core of team performance, yet it remains one of the most elusive phenomena to define and quantify \cite{marcelino2020collective}. Most analytical methods focus on isolated trajectories \cite{trident}, failing to capture underlying coordination as measurable data. This ball-centric bias persists even in state-of-the-art datasets like HockeyAI \cite{HockeyAI} and Club-Football \cite{gabor2025clubfootball}, which prioritize on-ball actions (e.g., puck-carrying or kicking) while overlooking critical off-ball spatial adjustments. This neglect of non-ball-carrying coordination is particularly limiting in badminton doubles \cite{ding2024estimation}. Unlike singles where individual execution remains foundational, doubles complexity is driven by reciprocal compensation \cite{Reciprocal_compensation}: a collaborative mechanism where players dynamically adjust positioning to cover spatial vulnerabilities. 
This persistent gap in sports data motivates us to redefine player dynamics through a formation-aware perspective, where a player’s movement is explicitly quantified in relation to their teammate.

Despite significant progress in individual sports analysis \cite{trident, van2022deepsportradar, sung2025player, deliege2021soccernet} and badminton singles \cite{wei2013predicting, spait, wang2023shuttleset}, two primary challenges of modeling doubles remain: (1) \textbf{the annotation of high-quality four-player spatial positions} and (2) \textbf{the lack of standardized and objective metrics to define and quantify team coordination}. As quantified in Table 1, existing sports datasets exhibit critical gaps across three dimensions: annotation precision, multi-agent dynamics, and tactical depth. While large-scale repositories like SoccerNet-v2 \cite{deliege2021soccernet} provide extensive event-level labels, they lack the stroke-level spatial context necessary for granular tactical analysis. Conversely, specialized racket sports datasets such as THETIS \cite{gourgari2013thetis}, TTNet \cite{voeikov2020ttnet}, and ShuttleSet \cite{wang2023shuttleset} focus strictly on singles, omitting the multi-player tracking required for doubles coordination. Crucially, formation-aware features, which are essential for capturing tactical synergy, are entirely absent from all existing public resources. 

To bridge these gaps, we propose a new comprehensive dataset, namely DShuttleNet, to provide the first stroke-level doubles dataset with synchronized spatial positions for all four players, uniquely integrating formation-aware representations and diagnostic point-loss reasons. This framework shifts the analytical focus from isolated trajectories to collaborative formations, enabling a reproducible diagnosis of team coordination. Our key contributions are summarized as follows:

\begin{table}[t]
    \centering
    \small
    \setlength{\tabcolsep}{1.7pt}

    \caption{Comparative analysis of existing sports datasets and ours.}
    \label{tab:Dataset_comparison}
    \renewcommand{\arraystretch}{1.1}

    \begin{tabular}{@{} l cccccccc @{}} 
        \toprule
        \textbf{Dataset (Year Venue)} & \makecell[c]{\textbf{Data} \\ \textbf{Scale}} & \textbf{Level} & \makecell[c]{\textbf{Multi-} \\ \textbf{Players}} & \makecell[c]{\textbf{Shot} \\ \textbf{Type}} & \makecell[c]{\textbf{Player} \\ \textbf{Positions}} & \makecell[c]{\textbf{Formation} \\ \textbf{Aware}} & \makecell[c]{\textbf{Point Loss} \\ \textbf{Reason}} \\
        \midrule
        DeepSportradar (2022 ACMMM) & 1.6M & Event & \boldmath{$\checkmark$} & $\times$ & \boldmath{$\checkmark$} & $\times$ & $\times$ \\
        SoccerNet-v2 (2021 CVPR) & 110,458 & Action & \boldmath{$\checkmark$} & \boldmath{$\checkmark$} & \boldmath{$\checkmark$} & $\times$ & $\times$ \\
        THETIS (2013 CVPR) & 8,374 & Action & $\times$ & \boldmath{$\checkmark$} & \boldmath{$\checkmark$} & $\times$ & $\times$ \\
        TTNet (2020 CVPR) & 4,271 & Event & $\times$ & $\times$ & \boldmath{$\checkmark$} & $\times$ & $\times$ \\
        ShuttleSet (2023 KDD) & 43,156 & Stroke & $\times$ & \boldmath{$\checkmark$} & \boldmath{$\checkmark$} & $\times$ & \boldmath{$\checkmark$} \\
        \midrule
        \textbf{\DatasetName~ (ours)} & \textbf{54,141} & \textbf{Stroke} & \boldmath{$\checkmark$} & \boldmath{$\checkmark$} & \boldmath{$\checkmark$} & \boldmath{$\checkmark$} & \boldmath{$\checkmark$} \\
        \bottomrule
    \end{tabular}
\end{table}

\begin{itemize}
     \item \textbf{Constructing a High-Quality, Expert-Annotated Dataset for Causal Evaluation:} We present the first public stroke-level badminton doubles dataset, comprising 58 matches, 4,483 rallies, and 54,141 shots from top-tier international tournaments between 2022 and 2025. Unlike existing datasets that focus on individual trajectories, \DatasetName~provides synchronized spatial positions for all four players and uniquely incorporates expert-annotated "Loss Reasons" categorized into six mutually exclusive types. This structural depth enables the dataset to serve as a diagnostic tool for causal tactical evaluation rather than a purely descriptive movement log. Dataset is available at: https://doi.org/10.34740/kaggle/dsv/16097938.
     \item \textbf{Defining Geometry-driven and Formation-aware Representations:} We formally define two geometric metrics: Teammate-Relative Angle (TRA) and Vertical Shot Angle (VSA) to characterize team coordination and shot aggressiveness in the context of doubles. By integrating both angular features, we establish a data-driven framework that provides a deep understanding between team formations and tactical dynamics, effectively eliminating the ambiguity inherent in traditional experience-based interpretations.
    \item \textbf{Proposing a Quantitative Metric for Teammate Coordination:} To address the challenge of quantifying interpersonal synergy, we introduce two novel metrics. On the offensive side, we propose \metrixDoubleRelativefull~(\metrixDoubleRelativeshort), which measures the success rate of the non-hitting partner in executing the reciprocal compensation, which is often overlooked in existing studies. On the defensive side, motivated by the Voronoi diagram framework \cite{shuttlecockdecay}, we introduce \metrixDefcoordinationfull~(\metrixDefcoordinationshort), which quantifies lateral spatial coordination by modeling defensive responsibility zones, allowing for the identification of severe spatial ruptures that compromise team integrity.
    \item \textbf{Empirical Validation of Tactical Performance:} Through extensive analyses and experiments, we demonstrate the efficacy of the proposed features and metrics. We further reveal that existing methods may suffer from underfitting when failing to incorporate these underlying coordination-centric features, thereby justifying their critical importance. Consequently, we identify new research directions facilitated by the coordination-centric features provided in \DatasetName.
\end{itemize}

\vspace{-0.5em}
\section{\DatasetName: Tactical Coordination Dataset for Badminton Doubles}
\label{sec:dataset}
\vspace{-0.5em}
\subsection{Expert-Driven Annotated Process and Diagnostic Feature Extraction}\label{sec:basic_statistics}
\vspace{-0.5em}
\begin{figure}[t]
  \centering
  \begin{minipage}{0.55\textwidth}
    \centering
    \includegraphics[width=\linewidth]{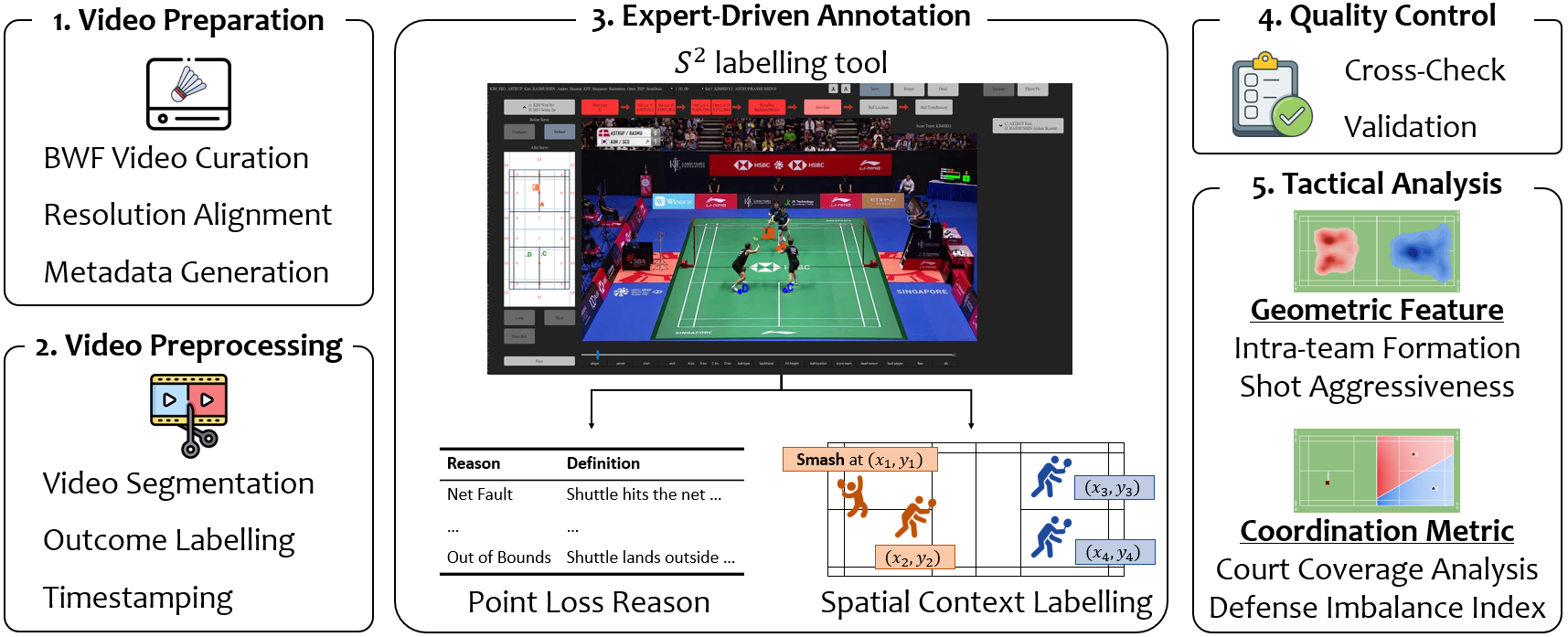}
    \caption{Expert-driven annotation pipeline.}
    \label{fig:annotation}
  \end{minipage}\hfill
  \begin{minipage}{0.45\textwidth}
    \centering
    \includegraphics[width=\linewidth]{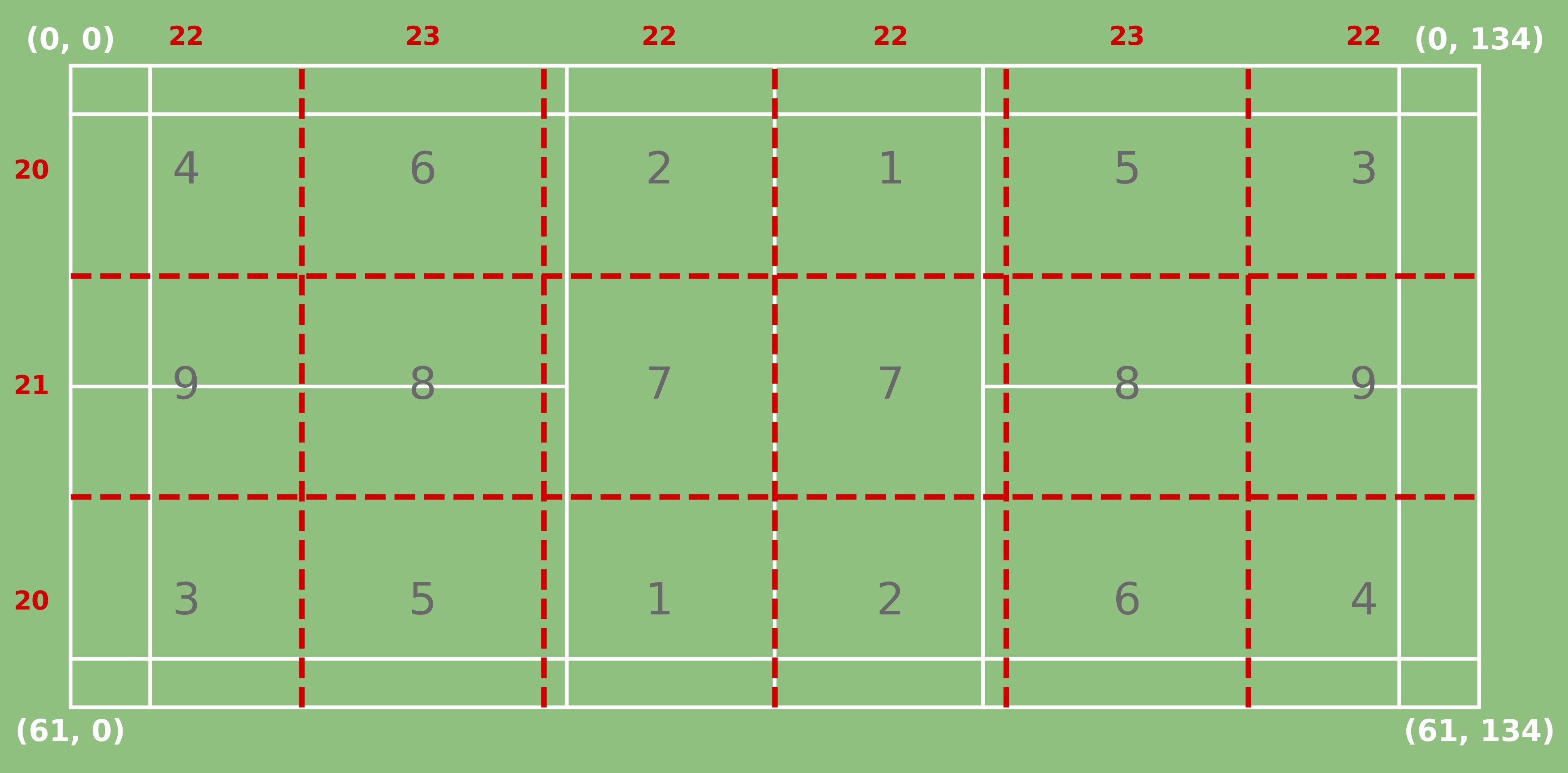}
    \caption{Uniform grid.}
    \label{fig:average_grid}
  \end{minipage}
\end{figure}
Table \ref{tab:Dataset_statistics} presents the basic statistics of \DatasetName, consisting of elite doubles matches spanning 2022 to 2025. While the overall average rally length is 12.08, both shot counts and rally lengths exhibit a distinct contrast between men's and women's doubles. Given that doubles competition emphasizes tactical coordination over individual shot variety, \DatasetName \ groups 18 shot types found in singles dataset \cite{wang2023shuttleset} into 9 primary types: Drop, Push, Smash, Drive, Lift, Net Shot, Clear, Long Serve, and Short Serve. Their definitions and statistical details are provided in Table \ref{tab:shot-details} and Table \ref{tab:Def_shot_type} in Appendix. 


\begin{table}[t]
    \centering
    \small
    \caption{Dataset statistics.}
    \label{tab:Dataset_statistics}
    \renewcommand{\arraystretch}{1.1} 
    \begin{tabular}{@{} l cccc @{}} 
        \toprule
        \textbf{Category} & \textbf{Matches} & \textbf{Rally} & \textbf{Shot Number} & \textbf{Avg. Rally Length} \\
        \midrule
        Men   & 24 & 1,841 & 15,344 & 8.33  \\
        Women & 24 & 1,843 & 31,737 & 17.22 \\
        Mixed & 10  & 799  & 7,060  & 8.84  \\
        \midrule
        \textbf{Total} & \textbf{58} & \textbf{4,483} & \textbf{54,141} & \textbf{12.08} \\
        \bottomrule
    \end{tabular}
\end{table}

Figure \ref{fig:annotation} illustrates our annotation pipeline. To guarantee professional-grade accuracy, we first follow existing protocols for video preprocessing \cite{huang2022s,wang2023shuttleset}, utilizing the shot-by-shot ($S^2$) labeling tool \cite{s2labeling} to extract stroke-level data. The court is mapped to a spatial reference frame from (0, 0) at the top-left to (61, 134) at the bottom-right. A uniform grid is applied to capture doubles-specific spatial dynamics, as illustrated in Figure~\ref{fig:average_grid}. The annotation process is then conducted by domain experts (national- and varsity-level doubles athletes and coaches), ensuring that every stroke is synchronized with: (1) \textbf{Spatial Context}: Frame-accurate (x, y) coordinates for all four players and the shuttlecock. (2) \textbf{Tactical Metadata}: Beyond player identities and shot types, we introduce \textbf{Point Loss Reasons} as a novel diagnostic feature, as shown in Table \ref{tab:point_loss_reasons}. Specifically, experts evaluate the outcome of each rally, categorizing the precise reason for point loss into mutually exclusive classes and identifying the specific player primarily responsible for the error. This unique addition enables the modeling of unforced errors and forced tactical breakdowns, providing a deeper layer of analysis not found in previous datasets. To ensure dataset reliability, domain experts randomly sampled and reviewed annotated rallies to verify the accuracy of spatial trajectories, scoring logic, and stroke-level records. Technical details are provided in Appendix \ref{sec:Detail Annotation}.

\begin{table}[ht]
    \centering
    \caption{Categorization and definitions of point loss reasons.}
    \label{tab:point_loss_reasons}
    \begin{tabular}{lp{0.6\linewidth}}
        \toprule
        \textbf{Point Loss Reason} & \textbf{Definition} \\
        \midrule
        Grounded within Boundaries & The shuttlecock lands within the valid court boundaries without the defending team making racket contact. \\
        \addlinespace
        Judgment Error & The defending player intentionally leaves the shuttlecock, misjudging it as out, but the landing coordinates fall within the boundaries. \\
        \addlinespace
        Net Fault & The shuttlecock is struck by a player, but its trajectory terminates by hitting the net mesh, failing to cross over to the opponent's side. \\
        \addlinespace
        Failed Return & The player attempts a stroke, but the shuttlecock lands on their own side of the court without reaching the net. \\
        \addlinespace
        Out of Bounds & The shuttlecock crosses the net, but its terminal landing point falls outside the legal court boundaries. \\
        \addlinespace
        Faults & Other actions resulting in an immediate point loss according to BWF regulations (e.g., service faults, racket touching the net, or shuttlecock hitting a player's body). \\
        \bottomrule
    \end{tabular}
\end{table}

\begin{figure}[htbp]
    \centering
    \begin{subfigure}[b]{0.26\textwidth}
        \centering
        \includegraphics[width=\linewidth]{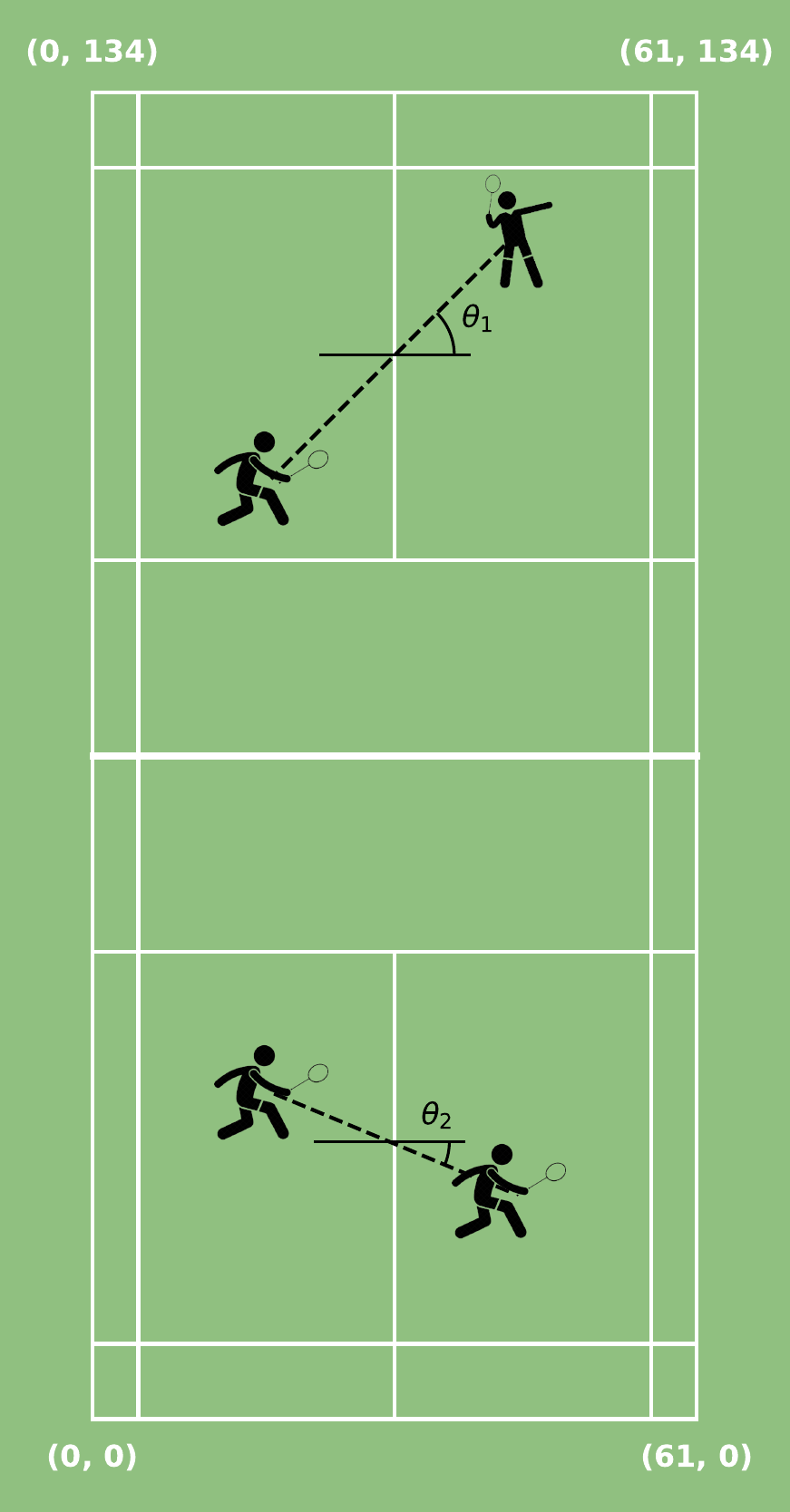}
        \caption{TRA example.}
        \label{fig:teammate_angle_example}
    \end{subfigure}\hfill
    \begin{subfigure}[b]{0.26\textwidth}
        \centering
        \includegraphics[width=\linewidth]{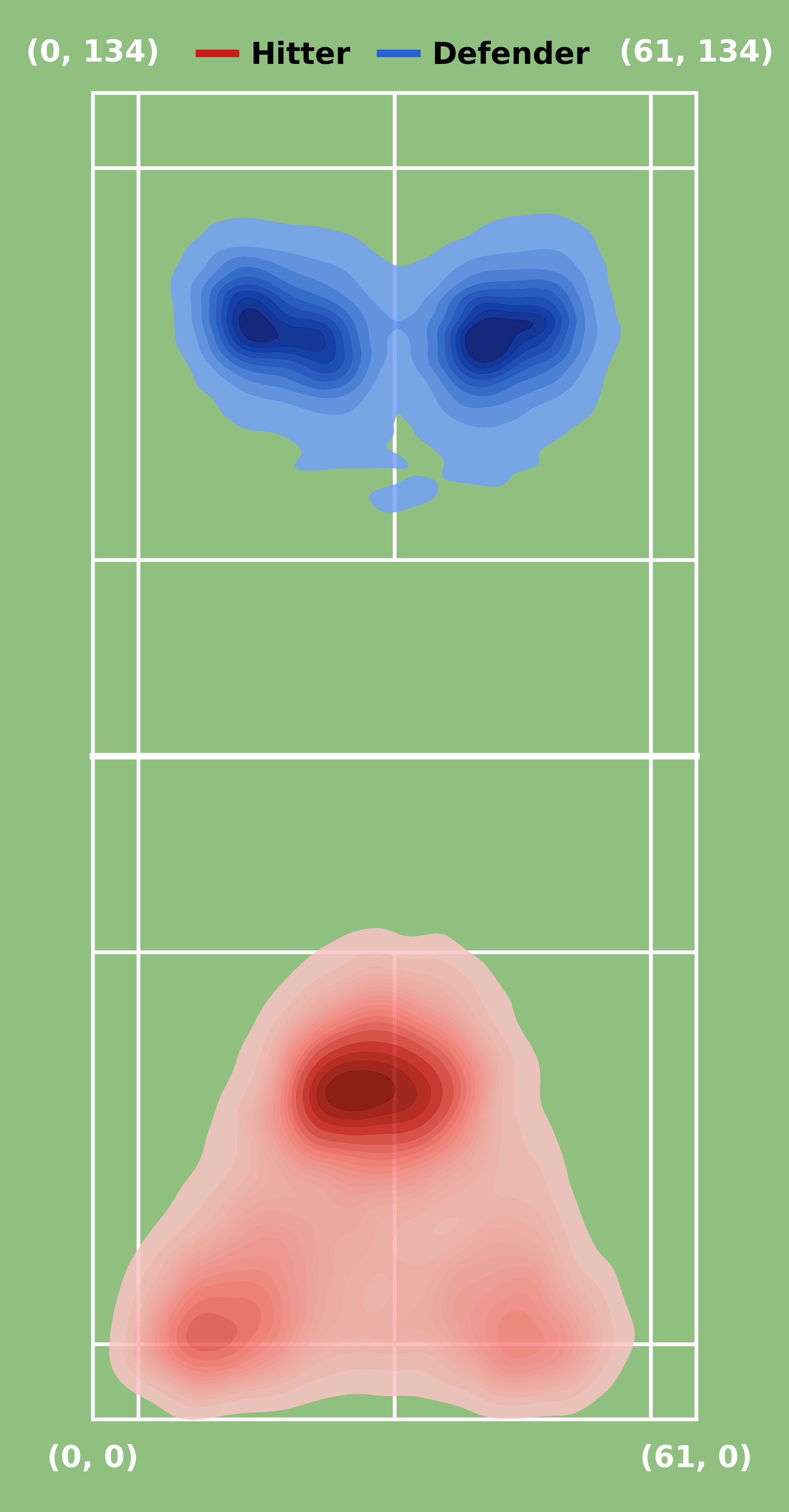}
        \caption{Smash position}
        \label{fig:smash_pos}
    \end{subfigure}\hfill
    \begin{subfigure}[b]{0.26\textwidth}
        \centering
        \includegraphics[width=\linewidth]{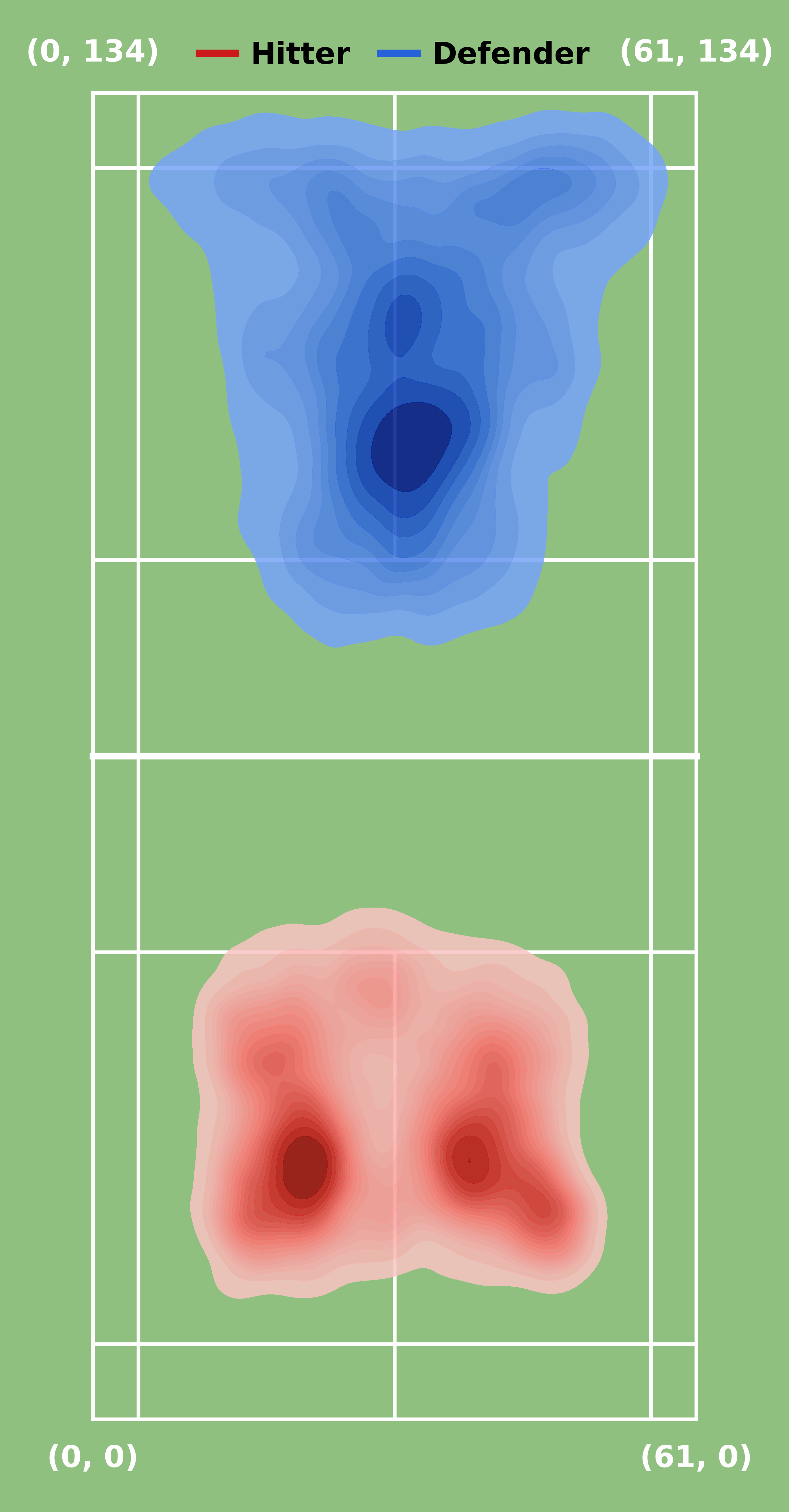}
        \caption{Lift position}
        \label{fig:pick_pos}
    \end{subfigure}
    \caption{Visualization of TRA and player distributions in smash and lift scenarios. (a) an angle-based feature for teammate positioning. (b) and (c) show the tactical formation selections: the attacking side adopts a front-back formation during a smash, while the defensive side uses a side-by-side formation after a lift to maximize court coverage.}
    \label{fig:angular_examples}
\end{figure}

\subsection{Geometry-driven Features}
\vspace{-0.5em}
\subsubsection{Teammate-relative Angle}
\label{sec:teammates_relative_angle}
\vspace{-0.8em}


To better understand team coordination, we introduce a novel geometric measurement, \textbf{Teammate-Relative Angle (TRA)}, $\theta\in [0^\circ, 90^\circ]$, defined as the minimum angle between the line connecting two teammates and the line parallel to the net, as shown in Figure \ref{fig:teammate_angle_example}. Lower angles ($\theta_2$) indicate a side-by-side defensive formation and higher angles ($\theta_1$) represent a front-back offensive alignment. While $\theta$ can be derived from raw $(x,y)$ trajectories, we explicitly model it to prevent tactical underrepresentation. Most existing trajectory models are designed for smooth movement data (see \cite{trident}), which is significantly different from doubles where players perform frequent sharp shuttle-runs. By providing $\theta$ as a primary feature, we ensure the model captures critical collaborative dynamics and formation transitions that are often lost in smoothed spatial position sequences.

\begin{figure}[htbp]
  \centering
  \begin{minipage}[t]{0.45\textwidth}
    \centering
    \includegraphics[width=\linewidth]{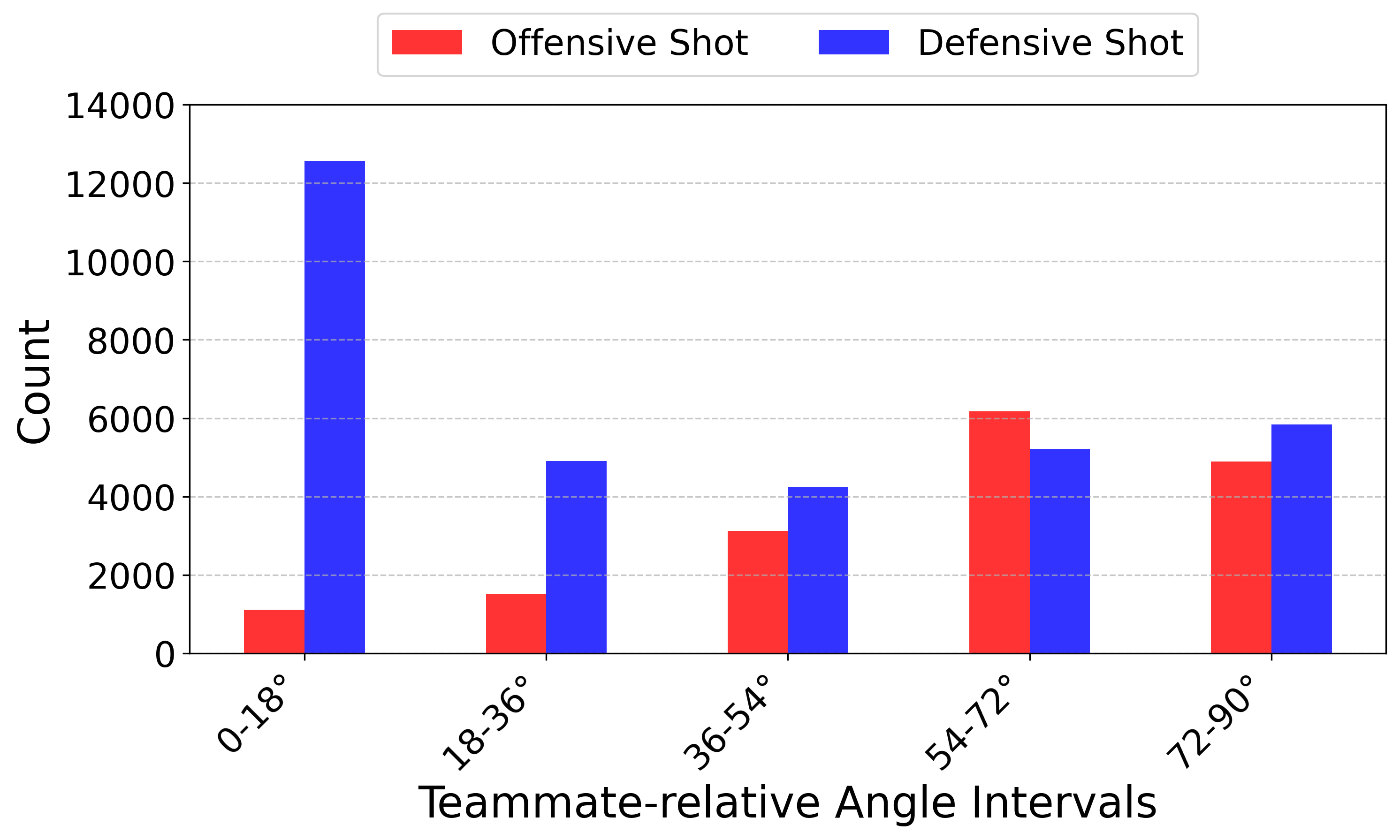}
    \caption{Distribution of shot types.}
    \label{fig:shot_type}
  \end{minipage}\hfill
  \begin{minipage}[t]{0.45\textwidth}
    \centering
    \includegraphics[width=\linewidth]{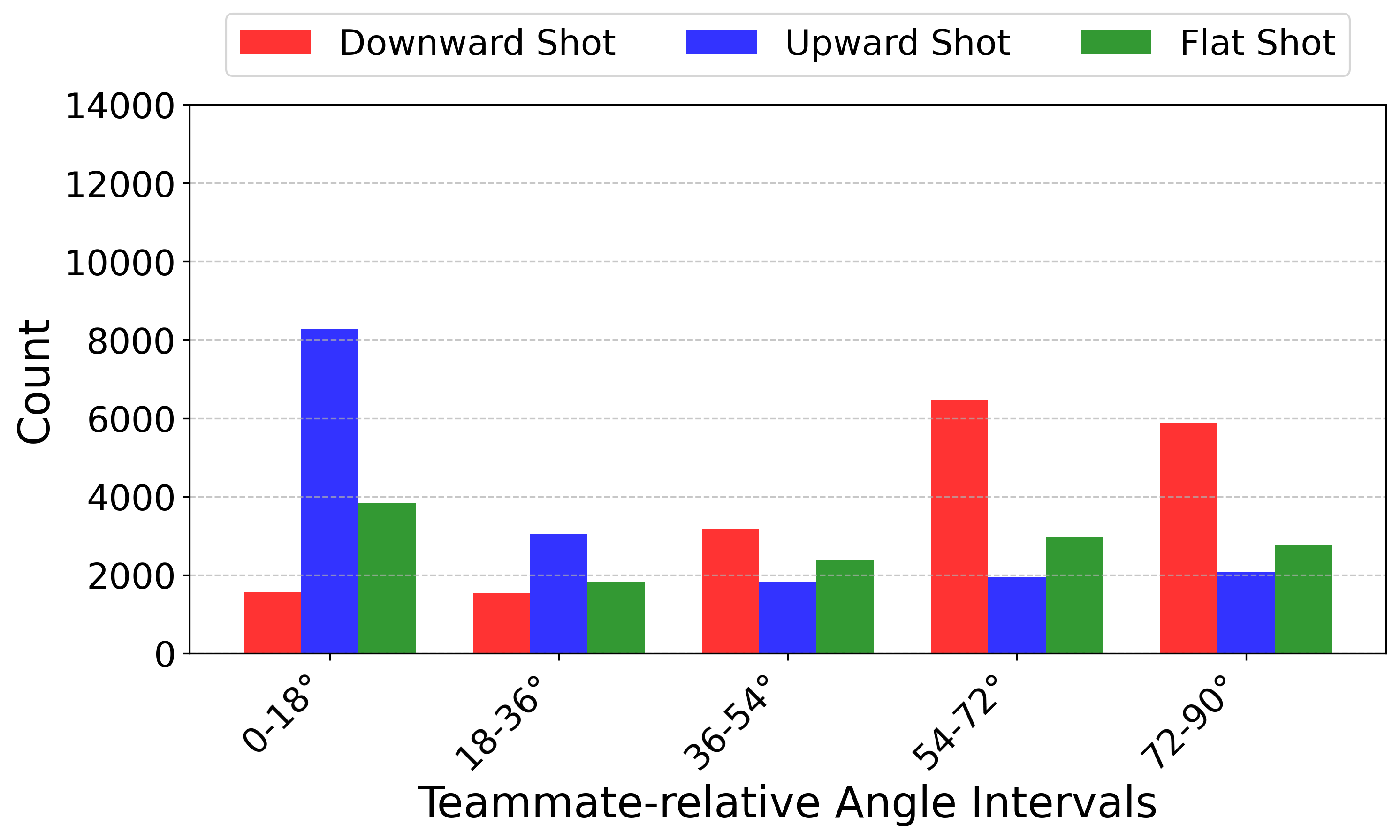}
    \caption{Distribution of VSA-based shot types.}
    \label{fig:angle-based}
  \end{minipage}
\end{figure}

Figure~\ref{fig:shot_type} illustrates the distribution of shot types with respect to given angular distributions. Offensive (i.e., Drop, Push, Smash) and defensive (i.e., Drive, Lift, Net Shot, Clear) shots exhibit a distinct trend: offensive shots predominantly occur at higher angles, reflecting an offensive formation, whereas defensive shots concentrate at lower angles. Consequently, this confirms a clear tactical separation based on teammates' angular values. 

\subsubsection{Vertical Shuttle Trajectory and Tactical Intent}
\label{sec:angle_based_shot_intent}
\vspace{-0.8em}



To quantify the tactical intent behind each stroke, \DatasetName~introduces \textbf{Vertical Shot Angle (VSA}) as a core categorical feature. Unlike existing ball-centric datasets that primarily record shuttle landing locations $(x, y)$, our dataset captures the verticality of each trajectory to distinguish between Downward, Flat, and Upward shots. This physical dimension serves as a critical bridge between raw movement and high-level tactical transitions.

As demonstrated in Figure 4, VSA exhibits a decisive correlation with team formations. Offensive downward shots (red) such as smashes and steep drops are concentrated in high-angle intervals ($54^{\circ}$- $90^{\circ}$). These shots represent an aggressive intent to terminate the rally, which typically triggers a front-and-back attacking formation. Conversely, defensive upward shots (blue bar), including lifts and clears, cluster in the lower angle ranges ($0^{\circ}$–$18^{\circ}$). These trajectories represent a "reset" intent, forcing the team to adopt a side-by-side defensive formation to maximize court coverage.  

The scientific value of providing VSA in \DatasetName~lies in its ability to eliminate the ambiguity of traditional shot labels. By explicitly encoding the vertical intent, we enable models to understand the causal flow of a rally: how a downward trajectory forces an immediate formation shift. This multi-dimensional representation allows for a fine-grained analysis of team coordination.  



\subsection{Spatial Dimension Analysis}
\begin{figure}[htbp]
  \centering
  \begin{minipage}{0.4\textwidth}
    \centering
    \includegraphics[width=\linewidth]{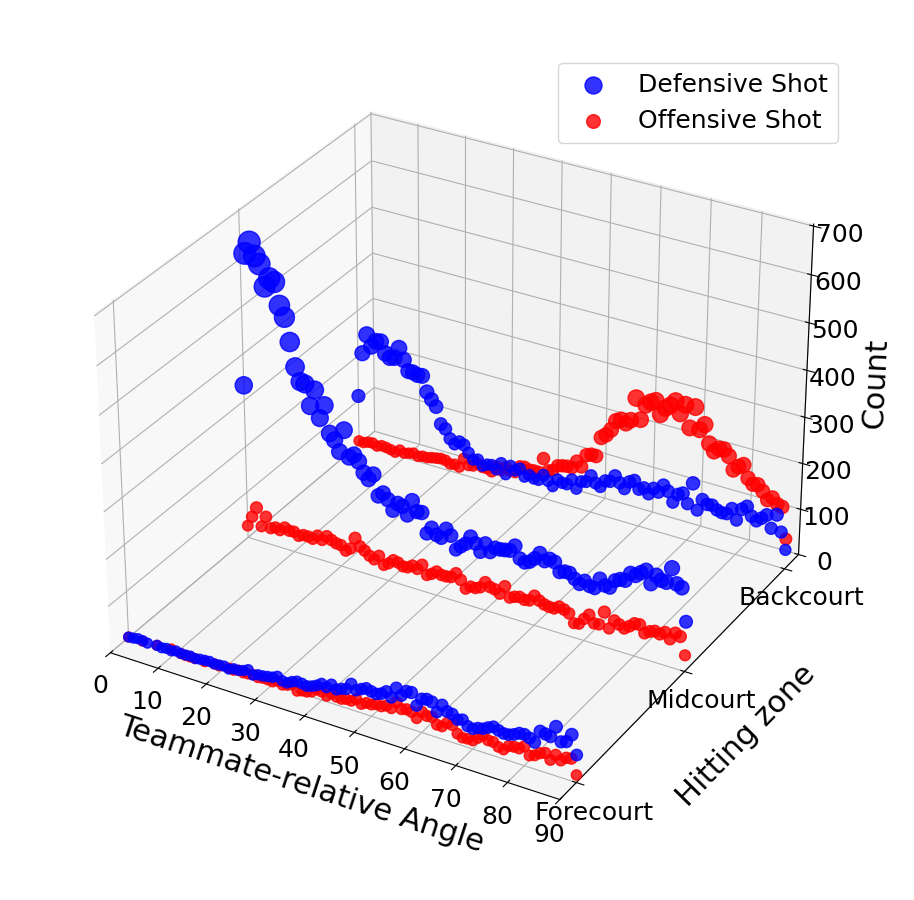}
    \caption{Distribution of shot types and spatial dimension.}
    \label{fig:3dplot_shottype}
  \end{minipage}\hfill
  \begin{minipage}{0.4\textwidth}
    \centering
    \includegraphics[width=\linewidth]{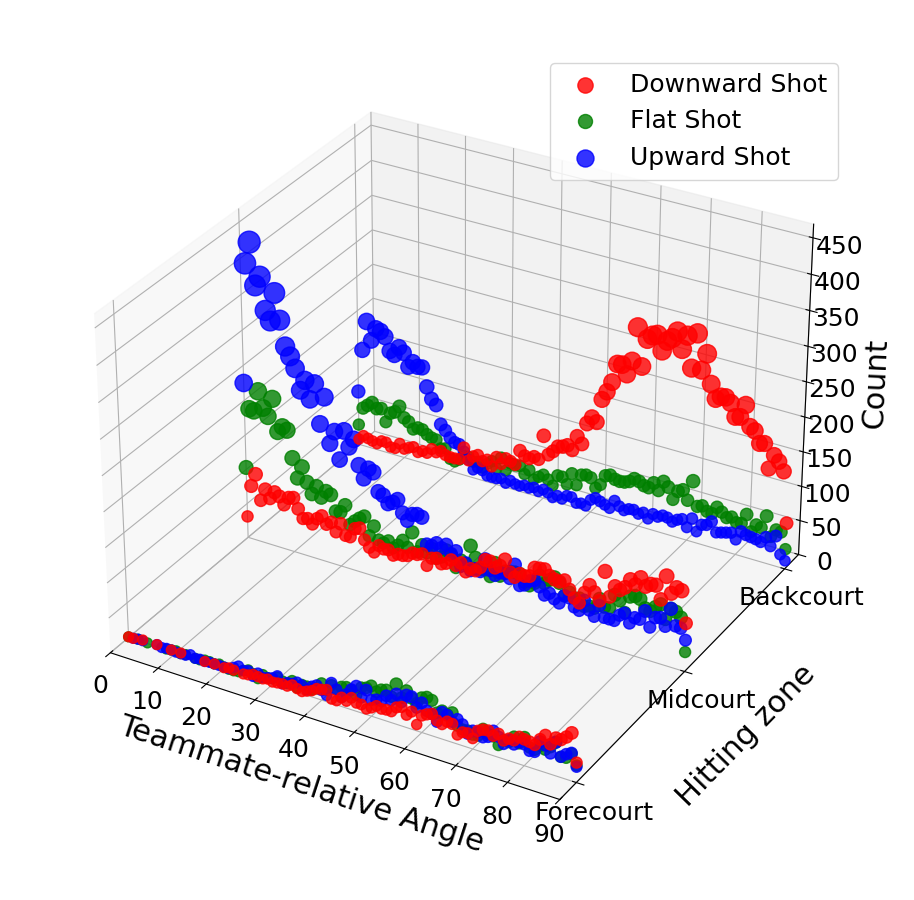}
    \caption{Distribution of VSA-based shot types and spatial dimension.}
    \label{fig:3dplot_angle_shot}
  \end{minipage}
\end{figure}

\vspace{-0.5em}
To demonstrate the unique analytical potential of this dataset, we integrate the spatial dimension with player-centric features through uniform grid subdivision. Figure \ref{fig:3dplot_shottype} showcases the synergy between team formation and shot selection at different locations. By leveraging the TRA, a clear tactical divide is quantified. Defensive shots (blue) are predominantly mapped to low-angle configurations within the Midcourt, while offensive shots (red) are concentrated at higher-angle formations in the Backcourt. This mapping confirms that our dataset effectively encodes the relationship between dynamic team positioning and game state.

Furthermore, Figure~\ref{fig:3dplot_angle_shot} uncovers tactical decisions with VSA under different locations and TRA. While upward shots (blue) predictably align with defensive patterns in the Midcourt, the inclusion of these intent-based angles allows for the identification of a critical transitional phase involving the struggle for attack initiative. Specifically, flat shots (green) reveal a tactical intent that is spatially similar to upward shots but functionally distinct in terms of rally control. Furthermore, the strong clustering of downward shots (red) at high teammate-relative angles in the Backcourt underscores the dataset's precision in representing aggressive scoring behaviors. These multi-dimensional features collectively provide a more granular representation of badminton tactics than previously available in existing datasets.

\vspace{-0.5em}
\section{Team Coordination-based Evaluation Metrics}
\label{sec:experiments}
\vspace{-0.5em}
While existing badminton analytics \cite{sung2024mocvae,wang2023shuttleset,lien2025shuttleflow} have extensively explored trajectory error, shot type classification, and hitter prediction, these metrics remain individual-centric and \textbf{fail to account for the non-hitting partner’s tactical role}. This paper identifies this limitation and introduces two novel metrics designed to quantify the effectiveness of teammate synchronization with respect to offensive and defensive scenarios: \metrixDoubleRelativefull~ (\metrixDoubleRelativeshort) and the \metrixDefcoordinationfull~ (\metrixDefcoordinationshort). These metrics shift the evaluation from individual-centric actions to the collaborative dynamics essential for doubles performance.

\begin{figure*}[t]
    \centering
    \begin{subfigure}[b]{0.26\textwidth}
        \centering
        \includegraphics[width=\linewidth]{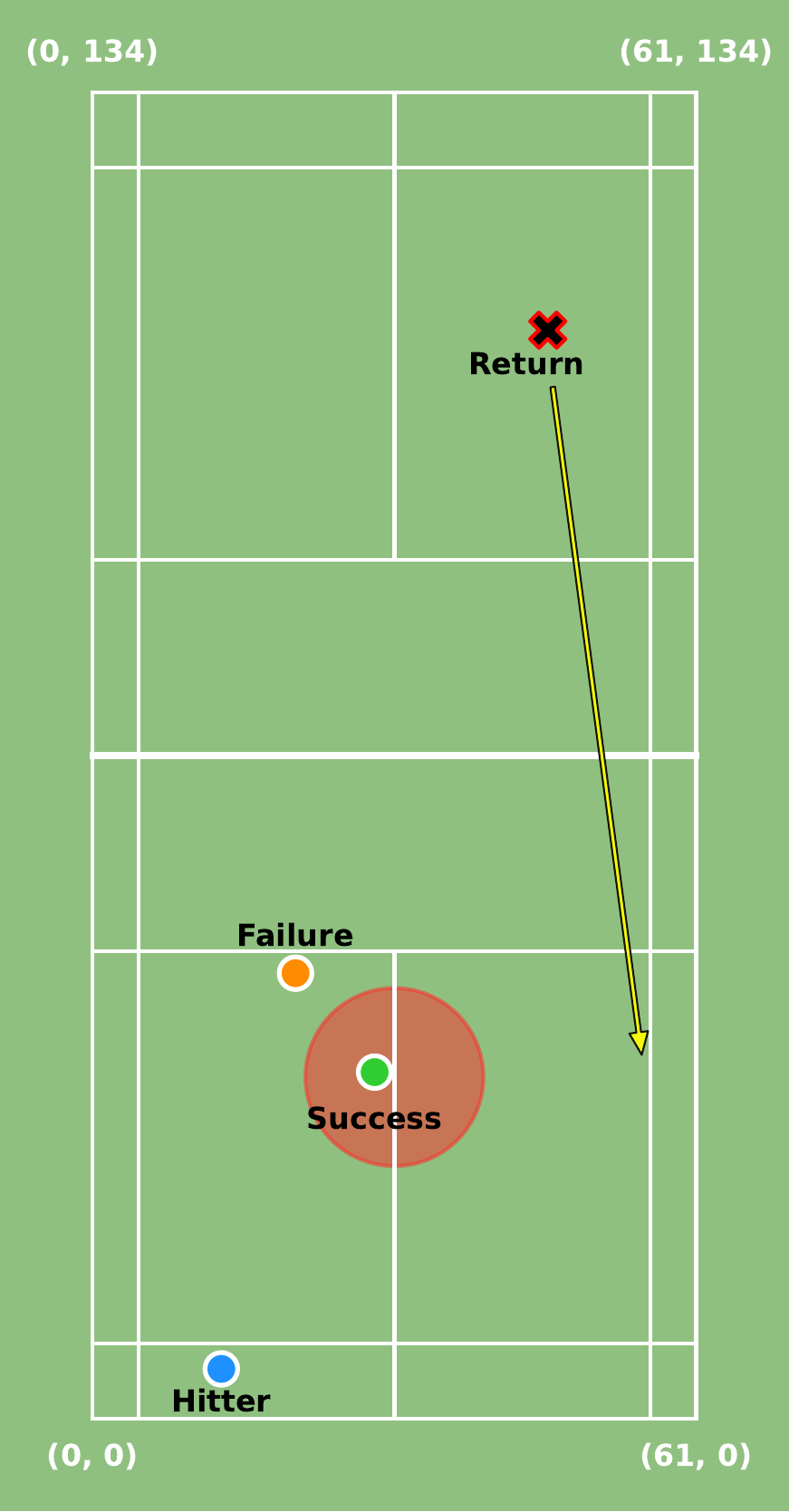}
        \caption{\metrixDoubleRelativeshort~}
        \label{fig:cce_schematic}
    \end{subfigure}\hfill
    \begin{subfigure}[b]{0.26\textwidth}
        \centering
        \includegraphics[width=\linewidth]{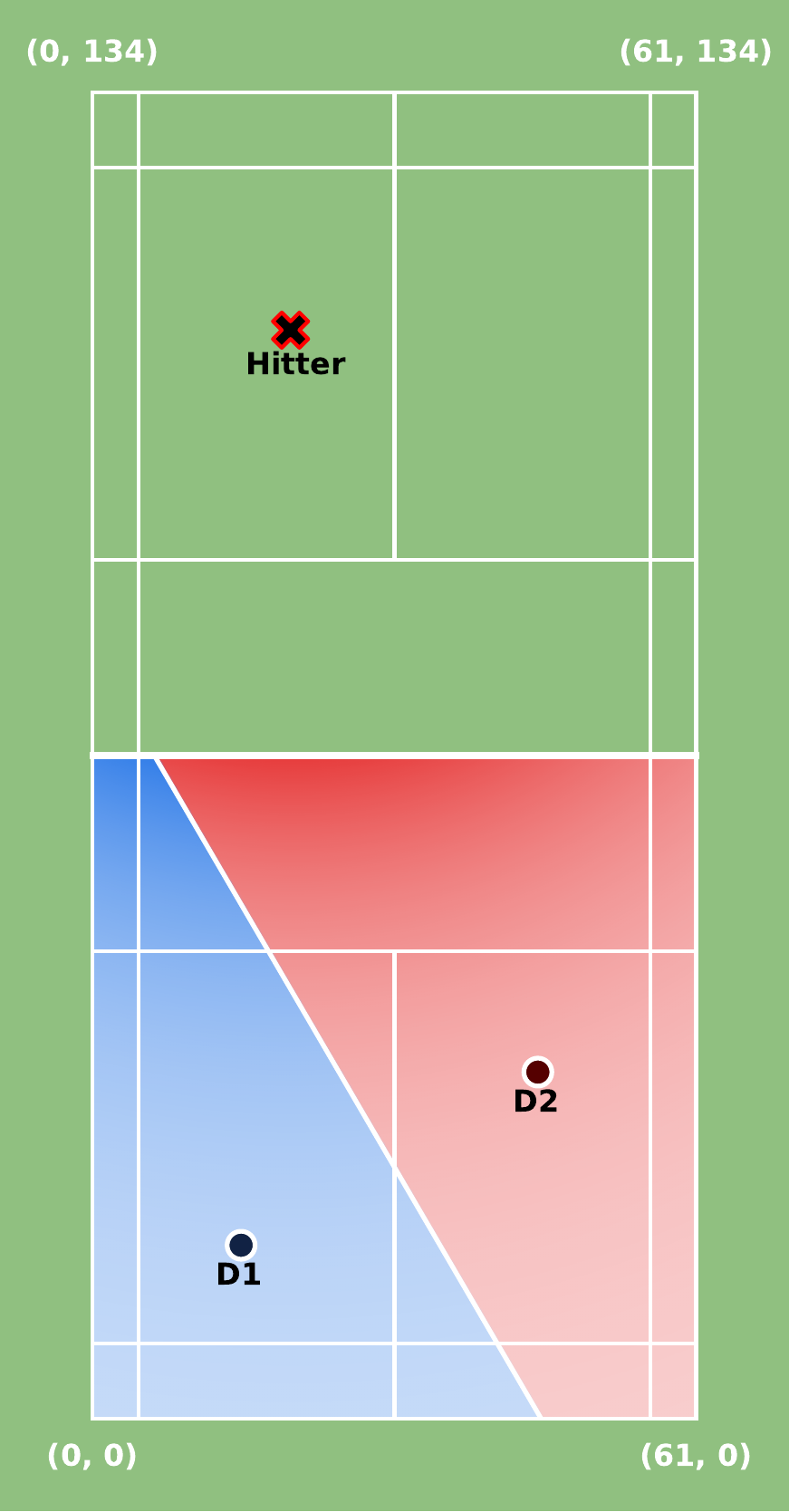}
        \caption{Unbalanced \metrixDefcoordinationshort}
        \label{fig:tlai_schematic_unbalance}
    \end{subfigure}\hfill
    \begin{subfigure}[b]{0.26\textwidth}
        \centering
        \includegraphics[width=\linewidth]{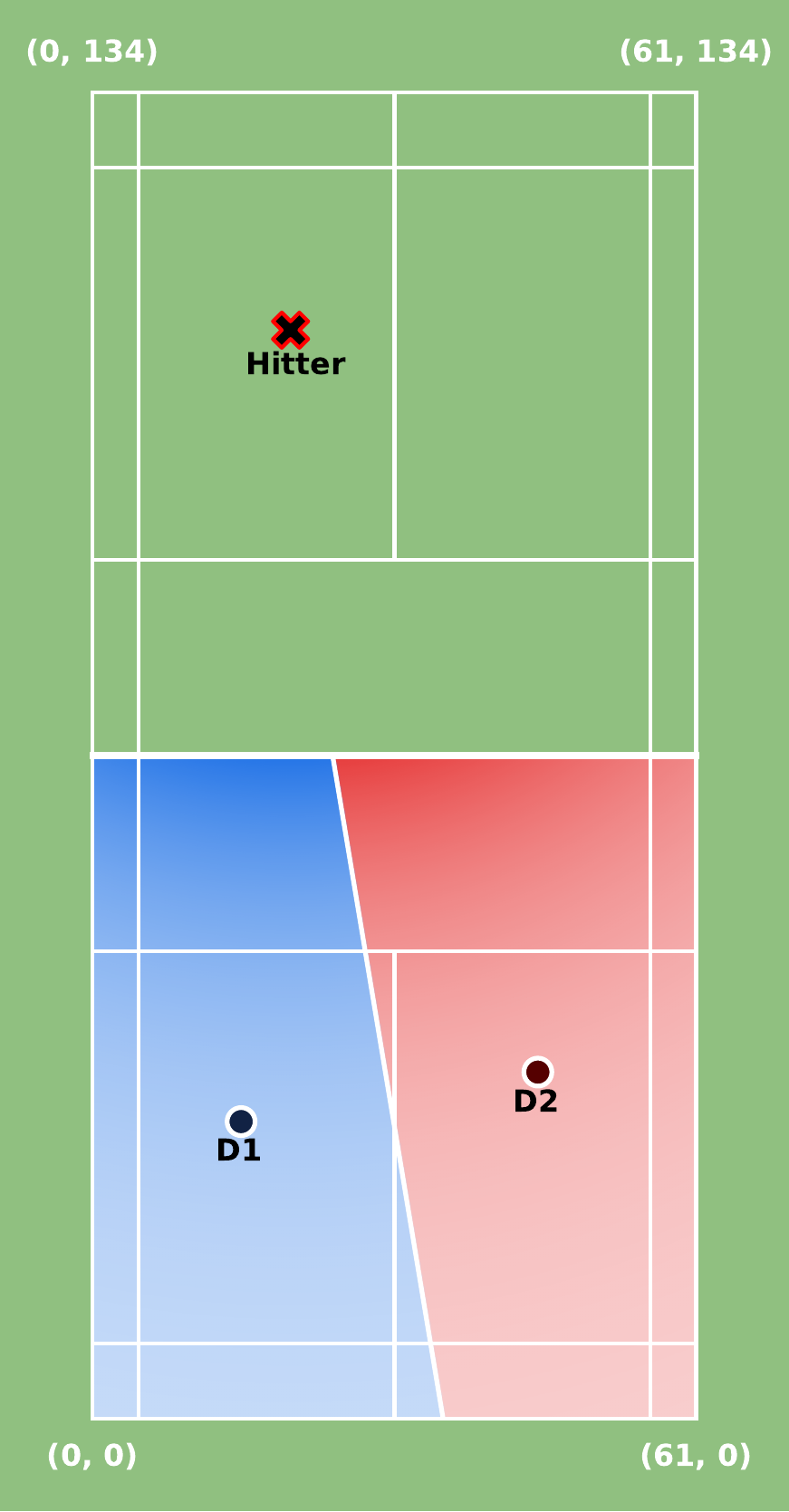}
        \caption{Balanced \metrixDefcoordinationshort}
        \label{fig:tlai_schematic_balance}
    \end{subfigure}
    
    \caption{Schematic representation of spatial metrics for double-match analysis. (a) \metrixDoubleRelativefull~(\metrixDoubleRelativeshort) illustrating the strategic intercept zone for balanced defensive coverage. (b) Threat Load Asymmetry Index (TLAI) with proportionate coverage. (c) Unbalanced TLAI where a minor shift by D1 significantly displaces the defensive boundary.}
    \label{fig:spatial_metrics_combined}
    \vspace{-0.5em}
\end{figure*}

\subsection{\metrixDoubleRelativefull~(\metrixDoubleRelativeshort)}
\vspace{-0.5em}
An effective attack relies not only on the hitter's aggressive strokes but also on the non-hitting partner's proactive spatial awareness, which is recognized as \textit{reciprocal compensation} \cite{Reciprocal_compensation}. Specifically, the partner must dynamically occupy the central defensive zone to mitigate spatial vulnerabilities created during the attack. Such center-focused coordinated positioning is crucial for maximizing the team's control area and minimizing opponents' scoring opportunities \cite{control_area}. As shown in Figure~\ref{fig:cce_schematic}, the strategic intercept zone is vital for controlling opponent returns. An out-of-position partner (Failure) may struggle to cover a distant return, leaving the other side of the court vulnerable. Conversely, a partner anchored within the zone (Success) maintains a balanced coverage. From the central pivot point, they can reach potential returns efficiently.

Consequently, to evaluate the non-hitting partner's proactive spatial readiness, we design \metrixDoubleRelativeshort, quantifying whether the team successfully covers a mid-court optimal pivot point $C$ as follows:
    $$\metrixDoubleRelativeshort =
    \begin{cases}
    1, & \text{if } \min\big(d(P_A, C), d(P_B, C)\big) \le R \\
    0, & \text{otherwise}
    \end{cases}$$
where $P_A,P_B$ denote the teammates and $d(\cdot, C)$ represents the Euclidean distance from either player to the pivot point. $C$ is spatially defined at coordinates $(30.5, 34.5)$, where is the central hub near the front service line (the T-junction). Geometrically, positioning at $C$ minimizes the maximum interception distance to potential counter-attack trajectories, such as net returns, blocks, and cross-court drives. The reciprocal compensation is deemed successful ($\metrixDoubleRelativeshort = 1$) if at least one player is positioned within a radial threshold $R$.


\subsection{\metrixDefcoordinationfull~(\metrixDefcoordinationshort)}
\vspace{-0.5em}
For the pair on the receiving end, the critical challenge is how teammates effectively balance and partition the court to prioritize defense against lethal threats. Since the reaction window narrows significantly near the opponent’s hitting point, we refine the Voronoi-based responsibility zones \cite{FONSECA20121652} by incorporating exponential velocity decay \cite{shuttlecockdecay}. This produces a \textit{kinematically-weighted area} that emphasizes urgent frontcourt burdens while discounting low-risk backcourt regions where increased flight time allows for easier recovery. Specifically, the court is first discretized into grids, where each grid $g$ assigned to the nearest defender to form their respective responsibility zones, $V_1$ and $V_2$. To prioritize the lethal zones where reaction time is minimal, we weight each grid based on the shuttlecock’s exponential velocity decay \cite{shuttlecockdecay}:
\begin{equation*}
    W(g) = 2^{-\frac{d(S, g)}{3.35}}, \quad TL_i = \sum_{g \in V_i} W(g)
\end{equation*}
where $W(g)$ represents the dynamic threat at grid $g$ relative to the attack origin $S$ and $TL_i$ is the cumulative threat load for defender $i$. In an ideal defensive formation, this burden should be distributed equitably between the two partners to avoid defensive collapse under high-pressure attacks. We thereby propose a novel metric, \metrixDefcoordinationshort, to evaluate the imbalance of this burden as follows:
\begin{equation*}
    TLAI = \frac{|TL_1 - TL_2|}{TL_1 + TL_2}
\end{equation*}
The index ranges from 0 to 1. A higher value indicates a significant difference in the threat-sharing, leading to an asymmetric loading where one defender is disproportionately pressured, as shown in Figure \ref{fig:tlai_schematic_unbalance}. In contrast, Figure \ref{fig:tlai_schematic_balance} demonstrates a more synchronized coordination example that effectively balances the defensive burden, resulting in a lower \metrixDefcoordinationshort~value. Consequently, \metrixDefcoordinationshort~serves as a quantitative metric for assessing the balance and coordination effectiveness at the receiving end. Please refer to Appendix~\ref{app:defensive_coordination} for more details.

\begin{figure}[t]
  \centering
  \begin{minipage}[t]{0.33\textwidth}
    \vspace{0pt} 
    \centering
    \includegraphics[width=\linewidth]{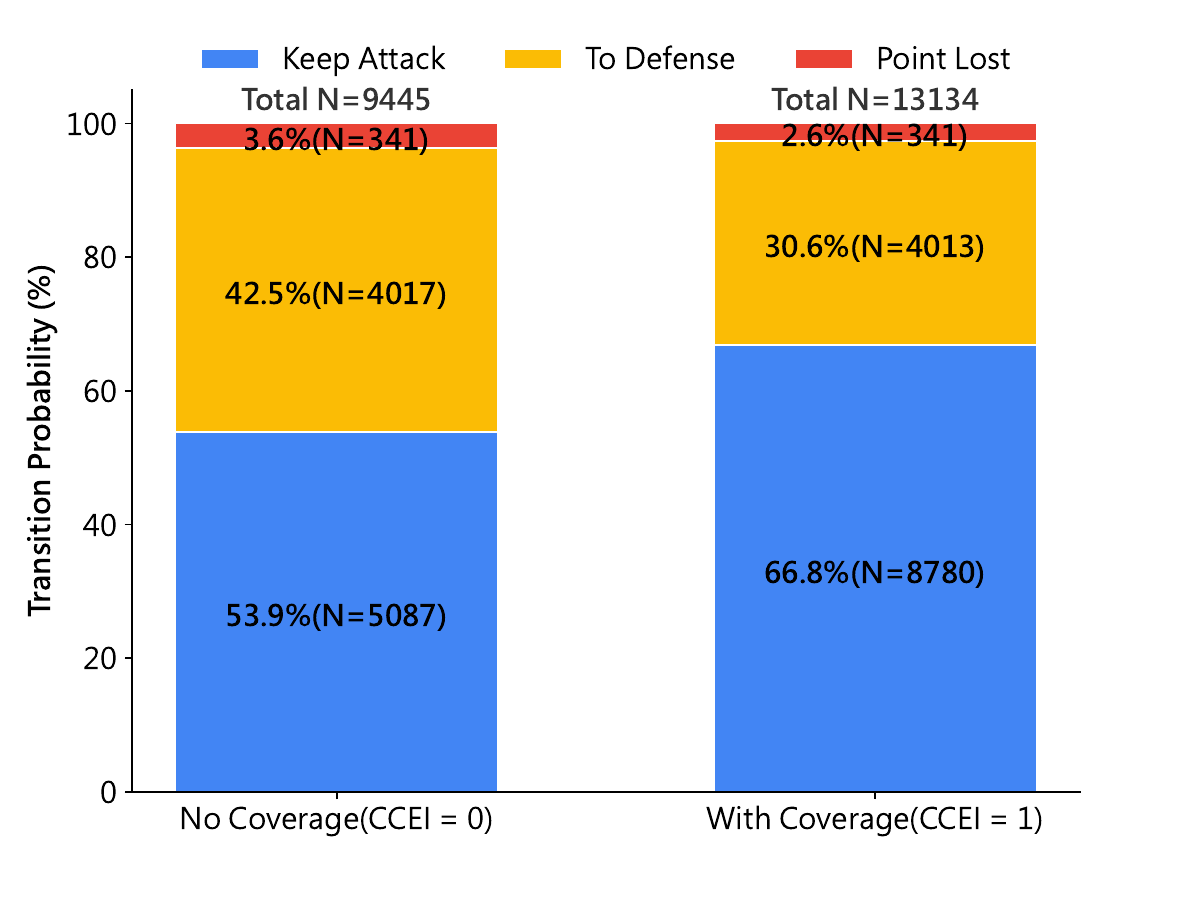}
    \caption{\metrixDoubleRelativeshort~transition probabilities at $R=9$~$(0.9\text{ m})$.}
    \label{fig:CCE_trans_prob_9}
  \end{minipage}\hfill
  \begin{minipage}[t]{0.33\textwidth}
    \vspace{0pt} 
    \centering
    \includegraphics[width=\linewidth]{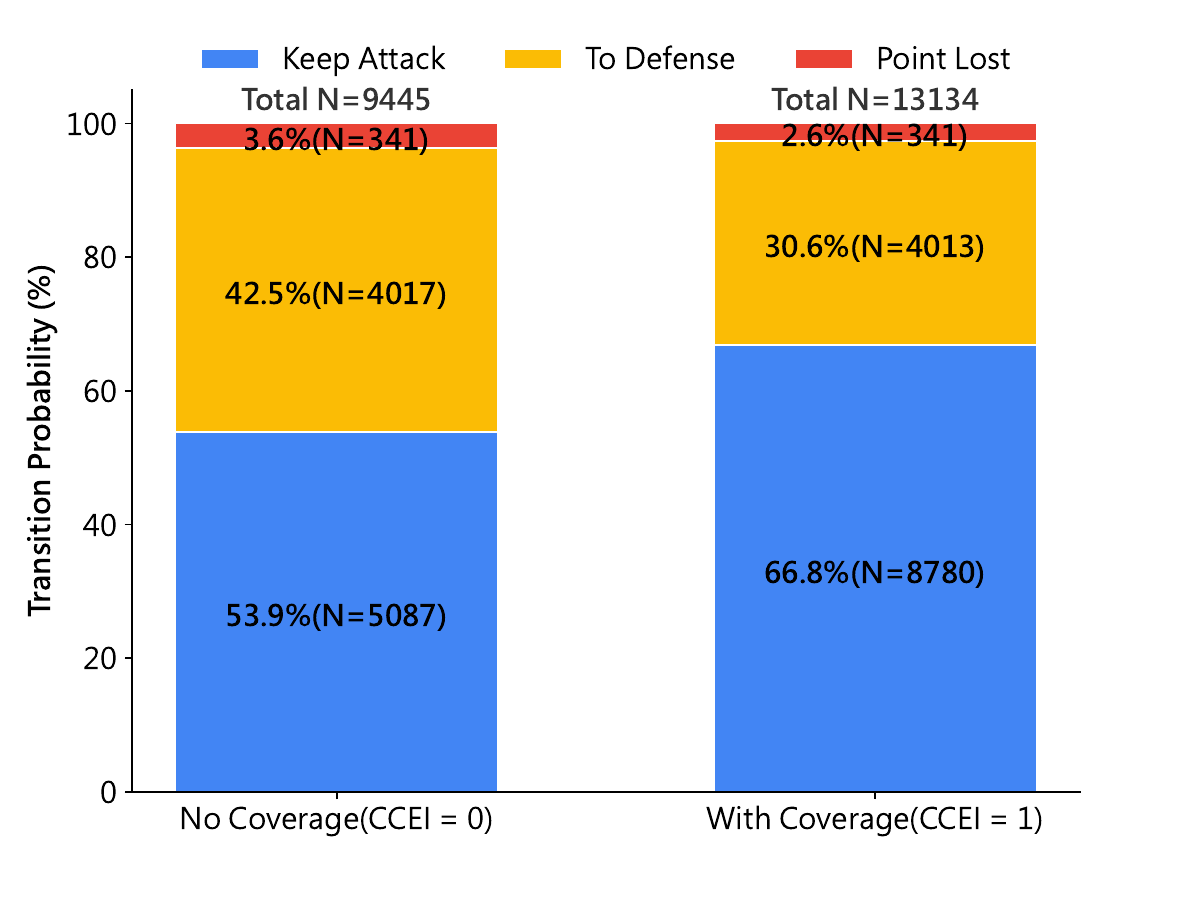}
    \caption{\metrixDoubleRelativeshort~transition probabilities at $R=15$$(1.5\text{ m})$.}
    \label{fig:CCE_trans_prob_15}
  \end{minipage}\hfill
  \begin{minipage}[t]{0.33\textwidth}
    \vspace{0pt} 
    \centering
    \includegraphics[width=\linewidth]{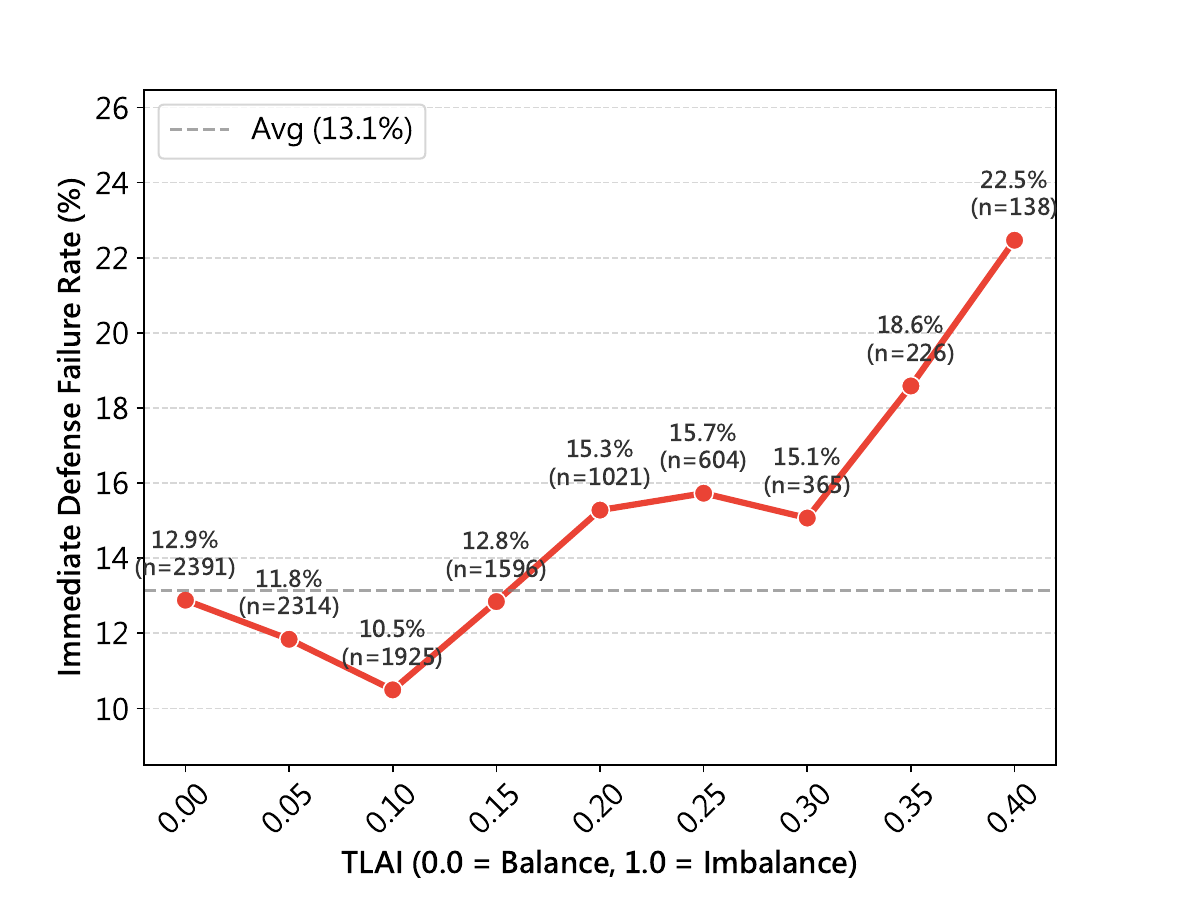}
    \caption{\metrixDefcoordinationshort~vs defensive failure rate.} 
    \label{fig:TLAI_threshold}
  \end{minipage}
\end{figure}

\subsection{Effectiveness Assessment of \metrixDoubleRelativeshort~and \metrixDefcoordinationshort}
\vspace{-0.5em}
\label{sec:effective_assessment_of_CCEI_and_TLAI}
To establish a causal link between spatial configuration and tactical outcomes, we apply a strict temporal boundary. A spatial configuration at stroke $N$ is classified as a ``failure'' if it directly results in a point loss or a forced error within the two subsequent strokes (i.e., $N+1$ or $N+2$). This microscopic temporal bounding isolates the immediate consequences of spatial imbalances from the noise of extended rallies. Both coordination metrics (\metrixDoubleRelativeshort~and \metrixDefcoordinationshort) are evaluated under this unified protocol to determine their impact on short-term rally survival.

\textbf{\metrixDoubleRelativefull.}
To evaluate preventive coordination under \attackform, we track the hitter team's exact tactical state at stroke $N+2$. Depending on rally continuity and the subsequent formation, the outcome is categorized into three mutually exclusive states: retaining the \attackform~(Keep Attack), being forced into a \defendform~(To Defense), or conceding the point (Point Lost). Figure~\ref{fig:CCE_trans_prob_9} illustrates the impact of mid-court coverage under strict conditions ($R=0.9\text{ m}$). While successful coverage ($\metrixDoubleRelativeshort = 1$) yields a 66.8\% attack continuation rate, failed coverage ($\metrixDoubleRelativeshort = 0$) not only reduces this rate to 53.9\% but also forces a transition to defense in 42.5\% of cases (up from 30.6\%).

Furthermore, the radius $R$ acts as a sensitivity threshold. As illustrated in Figure~\ref{fig:CCE_trans_prob_15}, a larger radius (e.g., $R=1.5\text{ m}$) establishes a looser positional constraint. Failing to achieve compensation under lenient conditions indicates a larger degree of positional deviation, leading to a more pronounced tactical disadvantage: the attack continuation rate drops to 34.4\%, while the transition to defense rises to 60.9\%. This comparison demonstrates that a strict $R$ evaluates general coordination proficiency, whereas a looser $R$ isolates critical positional breakdowns.

\textbf{\metrixDefcoordinationfull.}
To evaluate spatial coordination under \defendform, we analyze the relationship between the \metrixDefcoordinationshort~and the immediate defense failure rate against high-threat attacks (i.e., smashes and drops). A quadratic logistic regression model revealed a significant non-linear relationship (quadratic term: $coef = 0.0322$, $p = 0.008$; linear term: $p = 0.866$), indicating that the risk of point loss accelerates as spatial imbalance worsens.

As illustrated in Figure~\ref{fig:TLAI_threshold}, a critical threshold emerges at \metrixDefcoordinationshort ~$\approx 0.2$ (a 6:4 spatial load distribution). Below this threshold, the failure rate remains stable between 10.5\% and 12.9\%. However, once the imbalance exceeds 0.2, the failure rate escalates sharply, peaking at 22.5\% in extreme scenarios (\metrixDefcoordinationshort $= 0.4$). These findings demonstrate that while elite defenders can tolerate minor spatial inequalities, exceeding this threshold severely compresses the disadvantaged player's reaction time, sharply increasing the probability of conceding a direct kill or forced error.
\vspace{-0.5em}
\section{New Research Directions with \DatasetName}
\vspace{-0.5em}
The proposed \DatasetName~supports several downstream tasks that capture the unique tactical dynamics of professional badminton doubles. Unlike singles, doubles relies heavily on rapid rotation and spatial coordination between partners. By leveraging our novel geometric features (TRA, VSA) and evaluation metrics (\metrixDoubleRelativeshort, \metrixDefcoordinationshort), we identify four primary research directions enabled by this data:

\textbf{Coordination-aware Behavior Prediction.} While trajectory and shot-type prediction have been explored in previous studies \cite{sung2024mocvae,sung2025player}, these models often suffer from tactical unawareness by failing to account for teammate interactions. Our empirical analysis in Table \ref{tab:overall_comparison} reveals that ignoring the TRA leads to oversmoothed trajectory predictions, which fail to capture the rapid rotations and sharp directional shifts inherent in professional doubles. This lack of coordination-awareness results in a significant Formation MSE of 10–15 degrees. Similarly, without incorporating the VSA, shot-type prediction accuracy achieves only at 46-50\%, as the model cannot distinguish between aggressive downward attacks and defensive lifts. These results suggest that current benchmarks are representationally inadequate, highlighting the necessity of integrating geometric features to achieve high-fidelity tactical modeling. Please refer to Appendix \ref{apdx:baseline} for experimental setup details.

\begin{table}[t]
    \centering
    \caption{Experimental results on trajectory and shot type prediction.}
    \label{tab:overall_comparison}
    \begin{tabular}{lccccc}
        \toprule
        \multirow{2}{*}{\textbf{Model}} & \textbf{Trajectory} & \textbf{Angle} & \multicolumn{3}{c}{\textbf{Shot Type Prediction}} \\
        \cmidrule(lr){2-2} \cmidrule(lr){3-3} \cmidrule(lr){4-6}
        & \textbf{FDE}  & \textbf{MSE} & \textbf{F1-score}  & \textbf{Accuracy} & \textbf{Cross Entropy} \\
        \midrule
        MoCVAE   & 0.9716 & 15.0876 & 0.2584 & 0.4657 & 1.9732 \\
        TSCVAE   &\textbf{ 0.5240}  & \textbf{10.2222} & \textbf{0.2951} & \textbf{0.5041}  & \textbf{1.9357} \\
        \bottomrule
    \end{tabular}
\end{table}

\textbf{Action-Value Estimation (AVE).} Moving beyond individual performance, this task focuses on the systemic impact of each stroke on team coordination. Although AVE is widely utilized for player assessment \cite{action_value_survey,neurips_av,how_is_stroke}, existing frameworks fail to account for the interpersonal dynamics inherent in multi-player settings. By utilizing the TRA, AVE can determine whether a specific stroke successfully facilitates an offensive transition (front-back formation) or forces the team into a vulnerable defensive state. Crucially, from the perspective of the non-hitting partner, the inclusion of \metrixDoubleRelativeshort~facilitates an evaluation of \textit{off-ball contributions}, a dimension frequently overlooked in existing sports analytics. This enables a synergy-centric valuation framework, where a shot’s merit is adjudicated not only by its immediate outcome but by its capacity to optimize the partner’s subsequent spatial options and the team’s overall win probability.

\textbf{Defensive Coverage and Failure Analysis.} A hallmark of elite doubles is the contribution of the non-hitting partners, as their positioning directly determines the team's defensive resilience. This task incorporates \metrixDoubleRelativeshort~and \metrixDefcoordinationshort~into a learning-based framework designed to quantify the integrity of defensive structures. By utilizing dynamic heatmaps and threat contour lines to visualize the spatial distribution of court pressure, it is expected to pinpoint 'miscommunication zones' where neither partner effectively covers the incoming threat. This approach shifts the analytical focus toward the proactive movement of the non-hitting partner, identifying whether they effectively pivot to close gaps or if an imbalance in defensive load, indicated by high \metrixDefcoordinationshort, leads to a tactical breakdown. Ultimately, this provides a data-driven methodology for partner-specific performance profiling that accounts for the essential coordination in doubles.

\textbf{Match Intensity and Fatigue Modeling.} Physical degradation throughout a match exerts a profound impact on the final outcome, yet its influence on team-level coordination is often difficult to quantify. Our dataset and proposed metrics enable a systematic framework to analyze how this fatigue leads to the degradation of team synergy. Researchers can leverage our data to track how exhaustion manifests as a decline in \metrixDoubleRelativeshort~or an escalation in the \metrixDefcoordinationshort, both of which correlate significantly with an increased probability of losing points in later stages of a match. Furthermore, we observe that fatigue reduces the volatility of the TRA, indicating that players become relatively static and lose the capacity for rapid offensive-defensive transitions. By identifying these fatigue-induced error patterns, such as delayed rotations or uneven load distribution, professional coaches can gain data-driven insights to optimize stamina management and maintain tactical discipline during late-match scenarios.
\vspace{-0.5em}
\section{Related Work}
\label{sec:related_work}
\vspace{-0.5em}
\subsection{Feature Representation for Interaction Modeling}
\label{sec:feature_representation_interaction}
\vspace{-0.5em}
Traditional sports analysis research has focused on the actions of individual players. However, Wu et al. \cite{SportsHHI} have proven that modeling the interactions between players is crucial for understanding high-level strategies. In terms of spatial representation, early research relied on absolute trajectories. Hauri et al. \cite{NBAPrediction} utilized Long Short-Term Memory (LSTM) to process continuous coordinate sequences in basketball. To incorporate tactical semantics, Xia et al. \cite{VREN} adopted a discrete approach by dividing the court into zones to describe the positioning of the players and annotating specific event attributes (e.g, blocker ball touches) in volleyball. Furthermore, to capture complex environmental cues, Huang et al. \cite{STcontext} introduced Spatial-Temporal Context to assist in online action detection. Addressing the complex relationships between players, graph-based modeling approaches \cite{GNN} treat players as nodes to capture dynamic dependencies on the court. However, existing representations seldom clearly encode the relationship between players, which hinders the modeling of cooperation and role distribution on the court. This gap makes it difficult to understand high-level tactical coordination.

\subsection{Prediction in Racket Sports}
\label{sec:prediction_racket_sports}
\vspace{-0.5em}


Research in racket sports has evolved to address game-specific characteristics.
In table tennis, where high velocity and complex spin challenge vision systems, Wang et al.~\cite{wang2025novel} integrated 3D coordinates with a physics-driven module for accurate trajectory prediction.
In tennis, early Hawk-Eye trajectory models~\cite{wei2013predicting} have advanced to 3D reconstruction from monocular videos~\cite{ertner2024synthnet} and tactical quantification~\cite{li2025tennis}.
For single badminton, previous work has focused on predicting player movement~\cite{spait, wang2024offline} and shuttlecock landing points~\cite{spait, wang2024offline}, and has even extended to addressing the explainability of AI predictions~\cite{wang2025shuttleshap}. Recent work has also introduced dense shot captioning for understanding full-match badminton video~\cite{bmfd}, while our focus is on teammate spatial coordination.
In recent badminton doubles research, existing approaches have focused on predicting trajectories based on 2D coordinates~\cite{sung2025player,sung2024mocvae}, with some studies further adding body features for positioning advice~\cite{ding2024estimation}.
However, existing doubles models remain largely individual-centric and lack explicit indicators for measuring teammate cooperation, making it difficult to capture the spatial relationships and collaborative interactions between players.
    
\vspace{-0.5em}
\section{Conclusion and Limitation}
\label{sec:Conclusion and Limitation}
\vspace{-0.5em}
We present \DatasetName, the first expert-annotated dataset for causal evaluation in badminton doubles. By defining geometry-driven features (TRA, VSA) and novel coordination metrics (\metrixDoubleRelativeshort, \metrixDefcoordinationshort), we shift the focus from individual trajectories to interpersonal synergy. Our analysis justifies these coordination-centric representations as essential for preventing model underfitting and capturing elite tactical dynamics. Despite the high quality of the data set produced by our annotation approach, it also has a few limitations. First, manual labeling is still necessary to guaranty annotation quality, and the loss reason also requires human judgment. Second, all videos are from BWF Super Series matches featuring elite players, so the captured tactical patterns may not fully generalize to recreational or amateur players.

\newpage
\bibliographystyle{abbrvnat}
\bibliography{reference}

\newpage
\appendix
\section*{Appendix}



\section{Dataset Details}

\begin{figure}[htbp]
    \centering
    \begin{tikzpicture}[
        node distance=0.8cm,
        box/.style={draw, rectangle, rounded corners, align=center, minimum height=1cm, text width=0.9\textwidth, fill=blue!5, font=\small},
        arrow/.style={draw, -latex, thick}
    ]

    \node[box] (tournament) {\textbf{Tournament} \\ (e.g., Tournament ID, Event Name, Year, Level, Location)};

    \node[box, below=of tournament] (match) {\textbf{Match} \\ (e.g., Match ID, Round, Duration, Set Count, Winner Team, Loser Team)};

    \node[box, below=of match] (set) {\textbf{Set} \\ (e.g., Set ID, Set Score, Set Winner)};

    \node[box, below=of set] (rally) {\textbf{Rally} \\ (e.g., Rally ID, Current Score, Score team, Serve Player, \textbf{Loss Reason}, \textbf{Fault Player})};

    \node[box, below=of rally] (stroke) {\textbf{Stroke} \\ (e.g., Stroke ID, Players' Position, Hitter, Stroke Characteristics, \textbf{Relative Angle}, \textbf{Formation}, \textbf{CCEI
    }, \textbf{TLAI})};

    \draw[arrow] (tournament) -- (match);
    \draw[arrow] (match) -- (set);
    \draw[arrow] (set) -- (rally);
    \draw[arrow] (rally) -- (stroke);

    \end{tikzpicture}
    \caption{Hierarchical data structure of the proposed badminton dataset. The architecture mirrors the top-down progression of a real-world tournament. A \textbf{Tournament} contains multiple \textbf{Matches}, which are divided into \textbf{Sets}. Each set is broken down into individual \textbf{Rallies}. At the microscopic level, each rally consists of sequential \textbf{Strokes} (individual hits). For each record, this dataset contains every detailed information ranging from tournament contexts down to stroke-level spatial metrics.}
    \label{fig:data_hierarchy}
\end{figure}
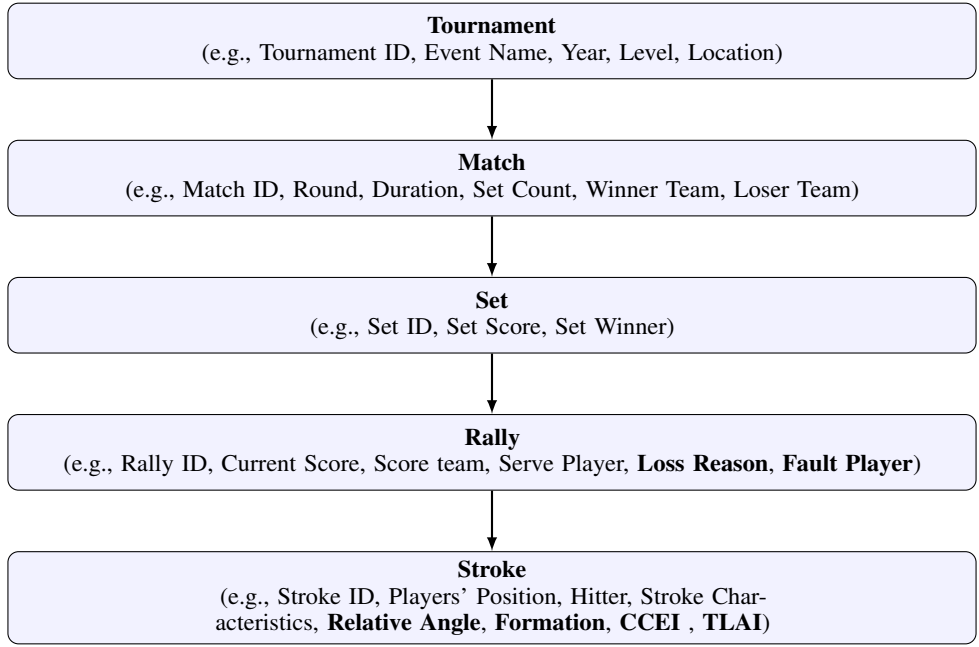

\begin{table}[h]
    \centering
    
    \setlength{\tabcolsep}{5pt} 
    
    \caption{Shot type details in the dataset.} 
    \label{tab:shot-details}
    
    \begin{tabular}{lcccc}
        \toprule 
        \textbf{Shot type} & \textbf{Men} & \textbf{Women} & \textbf{Mixed} & \textbf{Total} \\ 
        \midrule 
        Drop   & 534 (3.48\%)   & 2632 (8.29\%)  & 232 (3.29\%)   & 3398 (6.28\%)  \\ 
        Push        & 1550 (10.10\%) & 1929 (6.08\%)  & 626 (8.87\%)   & 4105 (7.58\%)  \\ 
        Smash       & 2297 (14.97\%) & 6165 (19.43\%) & 901 (12.76\%)  & 9363 (17.29\%) \\ 
        Drive        & 2950 (19.23\%) & 3803 (11.98\%) & 1141 (16.16\%)  & 7894 (14.58\%) \\ 
        Lift        & 2171 (14.15\%) & 7984 (25.16\%) & 919 (13.02\%)  & 11074 (20.45\%) \\ 
        Net shot    & 3836 (25.00\%) & 5111 (16.10\%) & 2319 (32.85\%) & 11266 (20.81\%) \\ 
        Clear       & 163 (1.06\%)   & 2257 (7.11\%)  & 122 (1.73\%)    & 2542 (4.70\%)  \\ 
        Long serve  & 237 (1.54\%)   & 222 (0.70\%)   & 136 (1.93\%)    & 595 (1.10\%)   \\ 
        Short serve & 1601 (10.43\%) & 1620 (5.10\%)  & 662 (9.38\%)   & 3883 (7.17\%)  \\ 
        \midrule
        \textbf{Total} & \textbf{15344} & \textbf{31737} & \textbf{7060} & \textbf{54141} \\ 
        \bottomrule 
    \end{tabular}
\end{table}
\begin{table}[ht]
    \centering
    \caption{Categorization and definitions of shot types.}
    \label{tab:Def_shot_type}
    \begin{tabular}{lp{0.8\linewidth}}
        \toprule
        \textbf{Shot Type} & \textbf{Definition}\\
        \midrule
        Drop & Hitting a lofted rear-court shuttle from above the head by altering the angle of contact between the racket and the shuttle. This changes the speed and trajectory, making the shuttle eventually drop near the net in a skillful stroke.  \newline \textbf{Note:} When the hitting speed is stronger, the shuttle drops further from the net but faster. When the hitting speed is weaker, the drop point is closer to the net (tight to the net) but slower.\\
        \addlinespace
        Push & A skillful technique of hitting a high shuttle located above the net with a fast stroke, pushing or killing the shuttle toward the opponent's court, causing it to drop rapidly downward. \newline \textbf{Note:} A stroke method that aims to be quick and oppress the opponent.  \\
        \addlinespace
        Smash &  Hitting a lofted shuttle from above the head by utilizing the racket's repulsion and bodily coordination (power transferred from the waist, back, and arm to the wrist). It is an extremely fast attacking technique hit from top to bottom into the opponent's court with strong and powerful speed. \newline \textbf{Note:} The main purpose is to destroy the opponent's defense and strive to score. \\
        \addlinespace
        Drive & A technique of hitting a mid-to-rear court shuttle at approximately waist-to-shoulder height back and forth with a flat trajectory and a relatively fast rhythm. \newline \textbf{Note:} A key hitting technique to suppress the opponent and transition from defense to offense with quick returns. \\
        \addlinespace
        Lift & A technique of hitting a front-court shuttle that is below net height, lifting it high and far forward from bottom to top, reaching the opponent's rear baseline. \newline \textbf{Note:} By lifting the shuttle high and far, you can buy yourself more and longer time and space to retreat and defend. \\
        \addlinespace
        Net Shot & A delicate technique of hitting a shuttle that is slightly below net height by relaxing the wrist to gently return the opponent's shot. By slicing the racket face across the shuttlecock's skirt, the shuttle crosses close to the net tape. \newline \textbf{Note:} The control of wrist movement and the angle of contact between the racket face and the shuttle at the moment of impact can produce a variety of different net shots to avoid direct attacks from the opponent. \\
        \addlinespace
        Clear & A technique of swinging the racket to return a lofted shuttle from your rear court over the opponent's head to the rear baseline. \newline \textbf{Note:} It is a skillful application of arm strength, wrist strength, stamina, and the correct contact point to buy time and seek attacking \\
         \addlinespace
        Long Serve &  Refers to a serve executed with a forehand or backhand motion, without violating the serving height limit (115 cm) and other service faults, hitting the shuttle over the receiver to land in the valid receiving area at the opponent's rear baseline. \newline \textbf{Note:} The longer the serve distance and the closer it is to the baseline of the valid receiving area, the more it can control the opponent's attacks and ambushes, making it advantageous for defense.\\
         \addlinespace
        Short Serve &  Refers to a serve executed with a forehand or backhand motion, without violating the serving height limit (115 cm) and other service faults, hitting the shuttle close to the net to drop quickly into the valid receiving area near the receiver's short service line. \newline \textbf{Note:} It can control the receiver's attacks and strive for the initiative on the next shot. \\
        \bottomrule
    \end{tabular}
\end{table}

\section{Additional Analysis}
\subsection{Spatial Distributions across Diverse Shot Types} \label{app:spatial_distribution}
\begin{figure*}[htbp]
    \centering
    \begin{subfigure}{0.32\textwidth}
        \includegraphics[width=\linewidth]{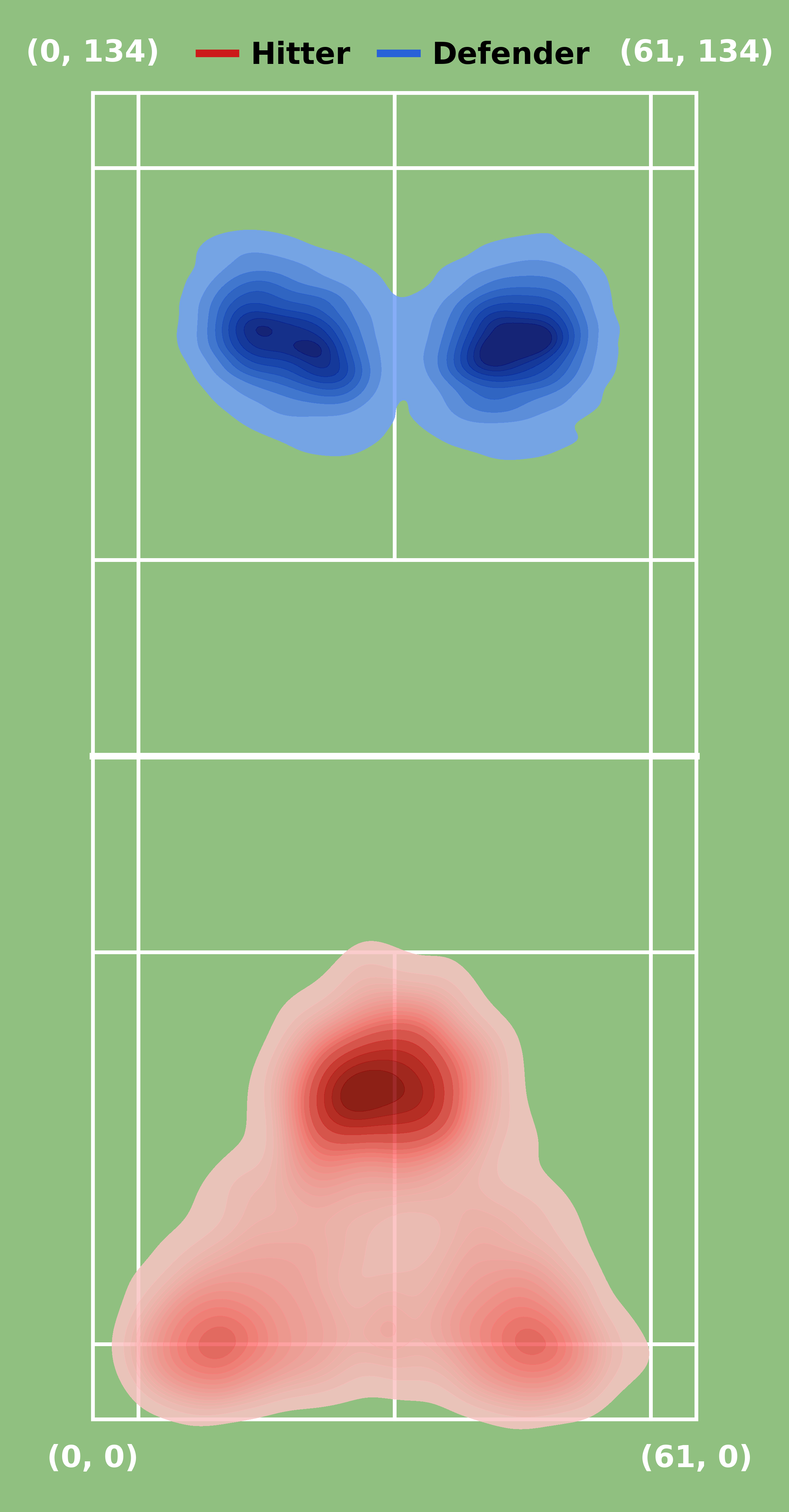}
        \caption{Drop position}
        \label{fig:Drop_pos}
    \end{subfigure}\hfill
    \begin{subfigure}{0.32\textwidth}
        \includegraphics[width=\linewidth]{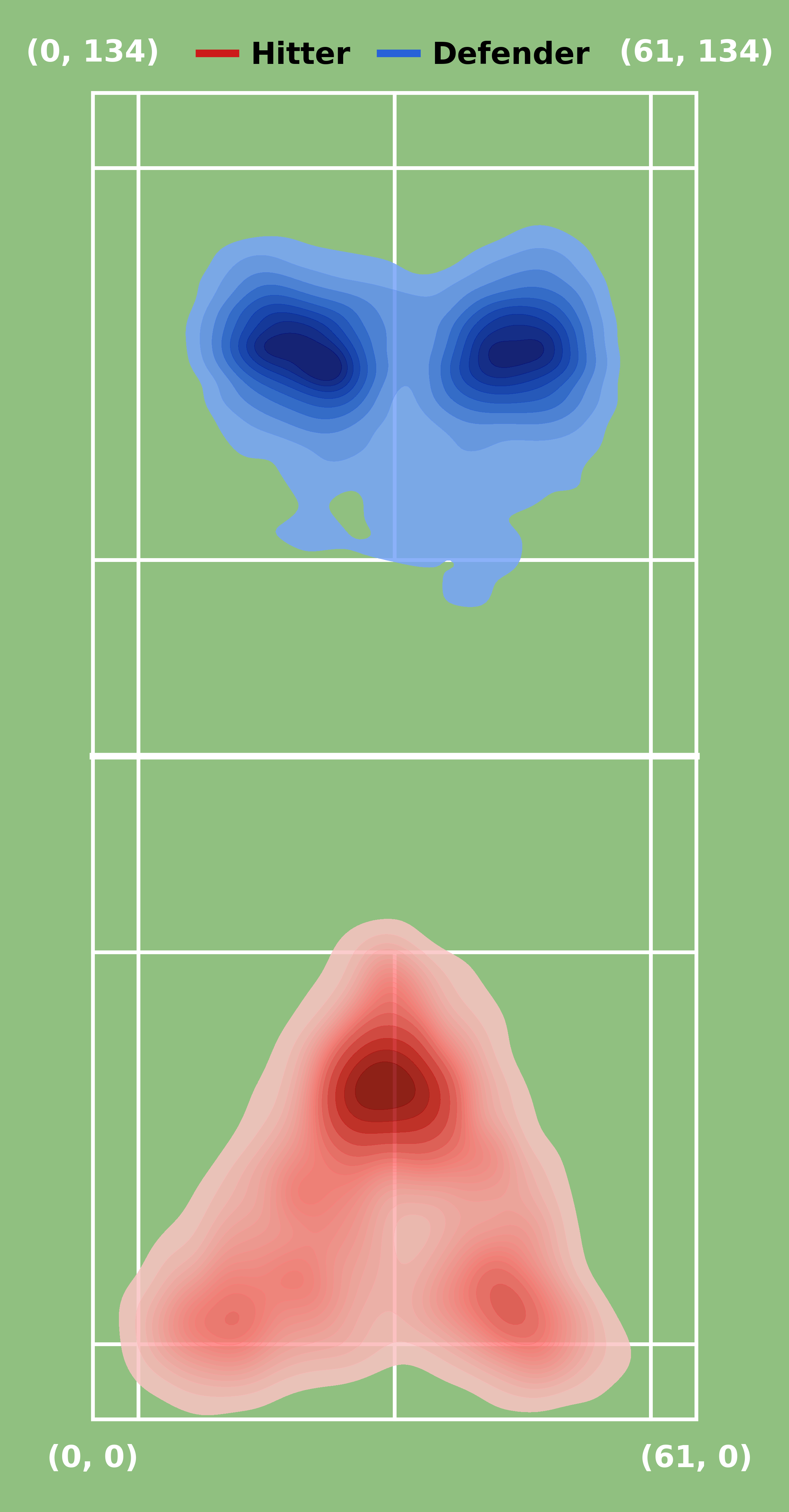}
        \caption{Clear position}
        \label{fig:clear_pos}
    \end{subfigure}\hfill
    \begin{subfigure}{0.32\textwidth}
        \includegraphics[width=\linewidth]{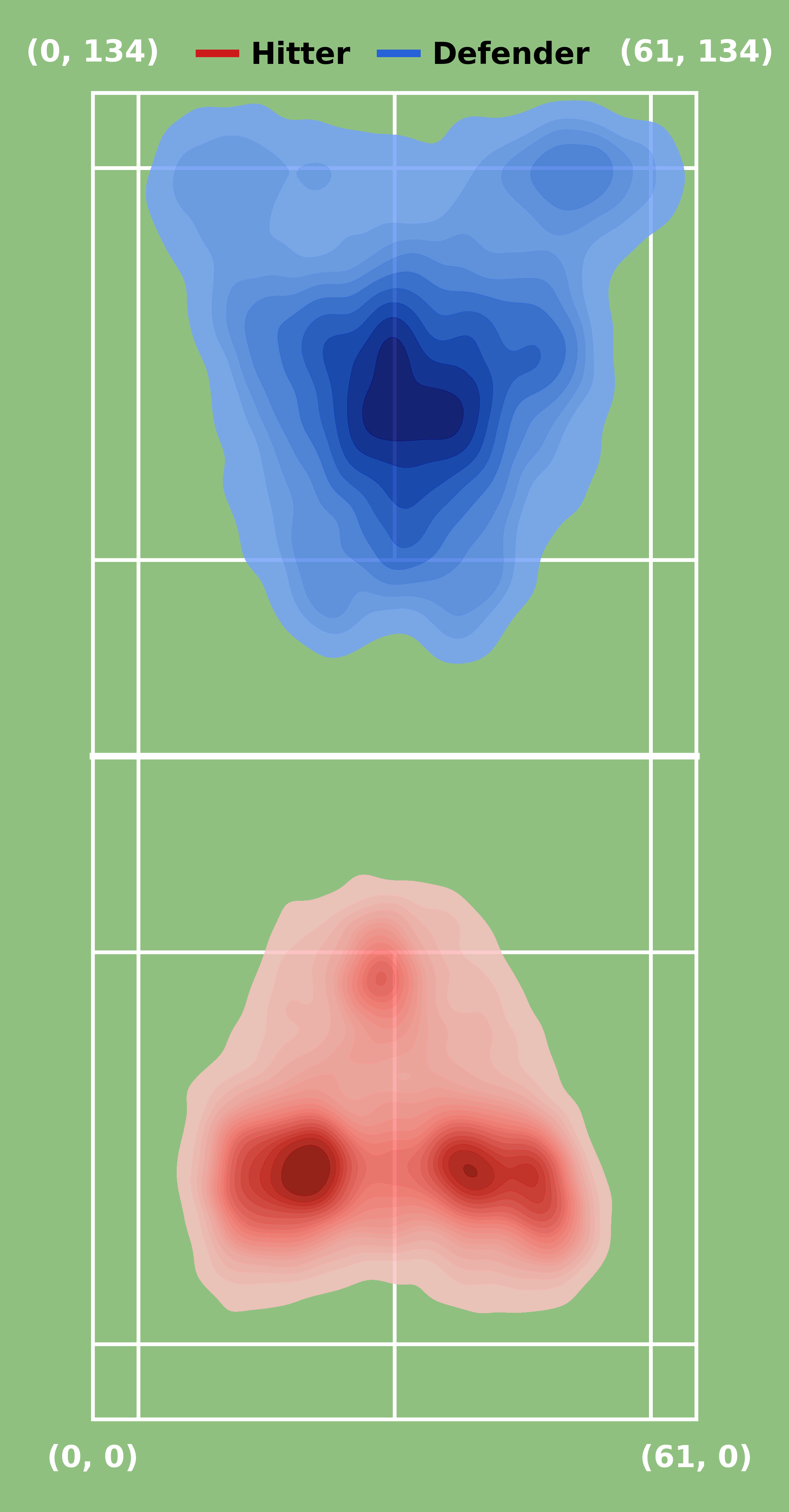}
        \caption{Drive position}
        \label{fig:drive_pos}
    \end{subfigure}
    
    \vspace{0.4cm} 
    
    \begin{subfigure}{0.32\textwidth}
        \includegraphics[width=\linewidth]{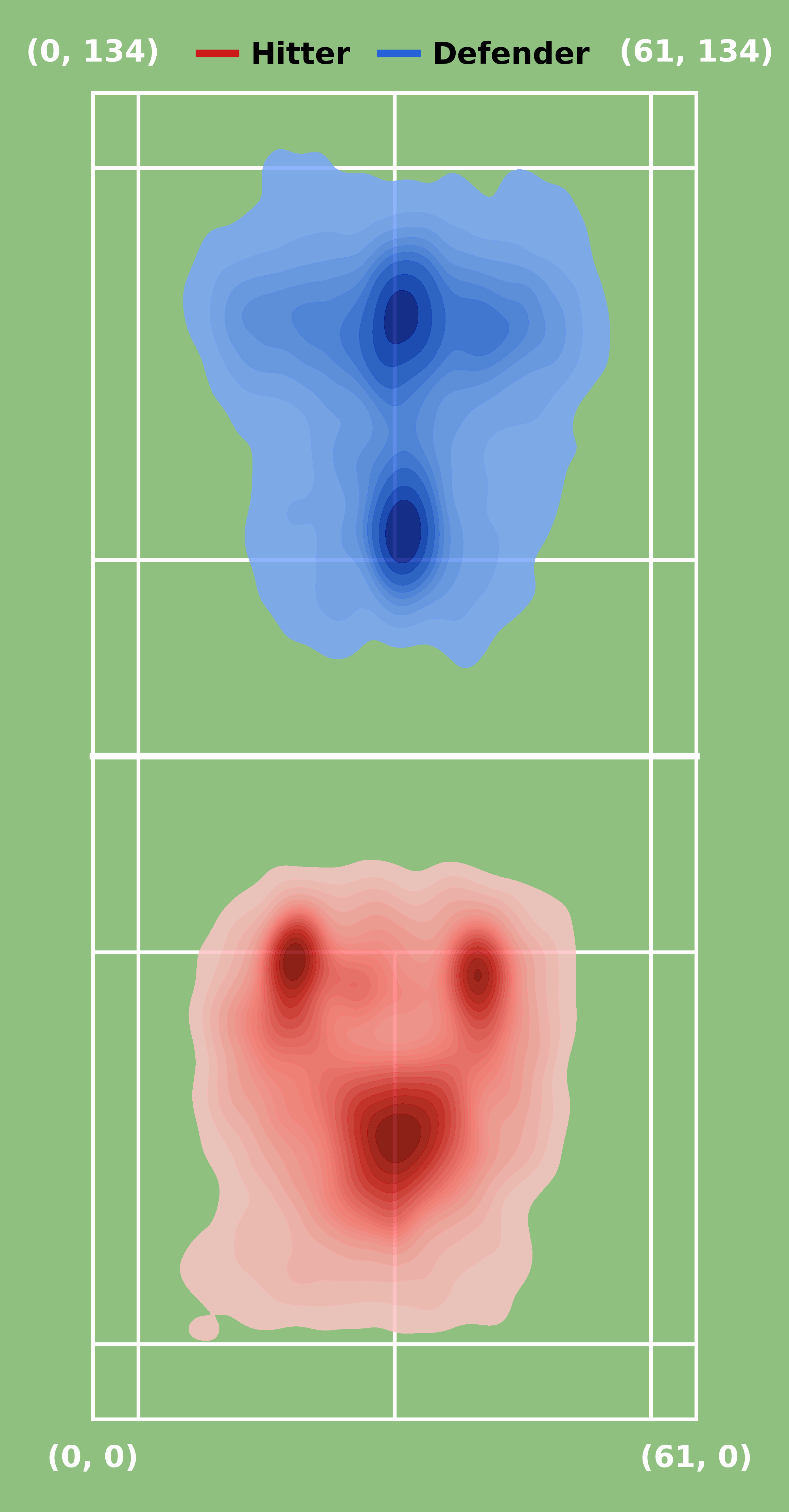}
        \caption{Push position}
        \label{fig:push_pos}
    \end{subfigure}
    \hspace{0.05\textwidth} 
    \begin{subfigure}{0.32\textwidth}
        \includegraphics[width=\linewidth]{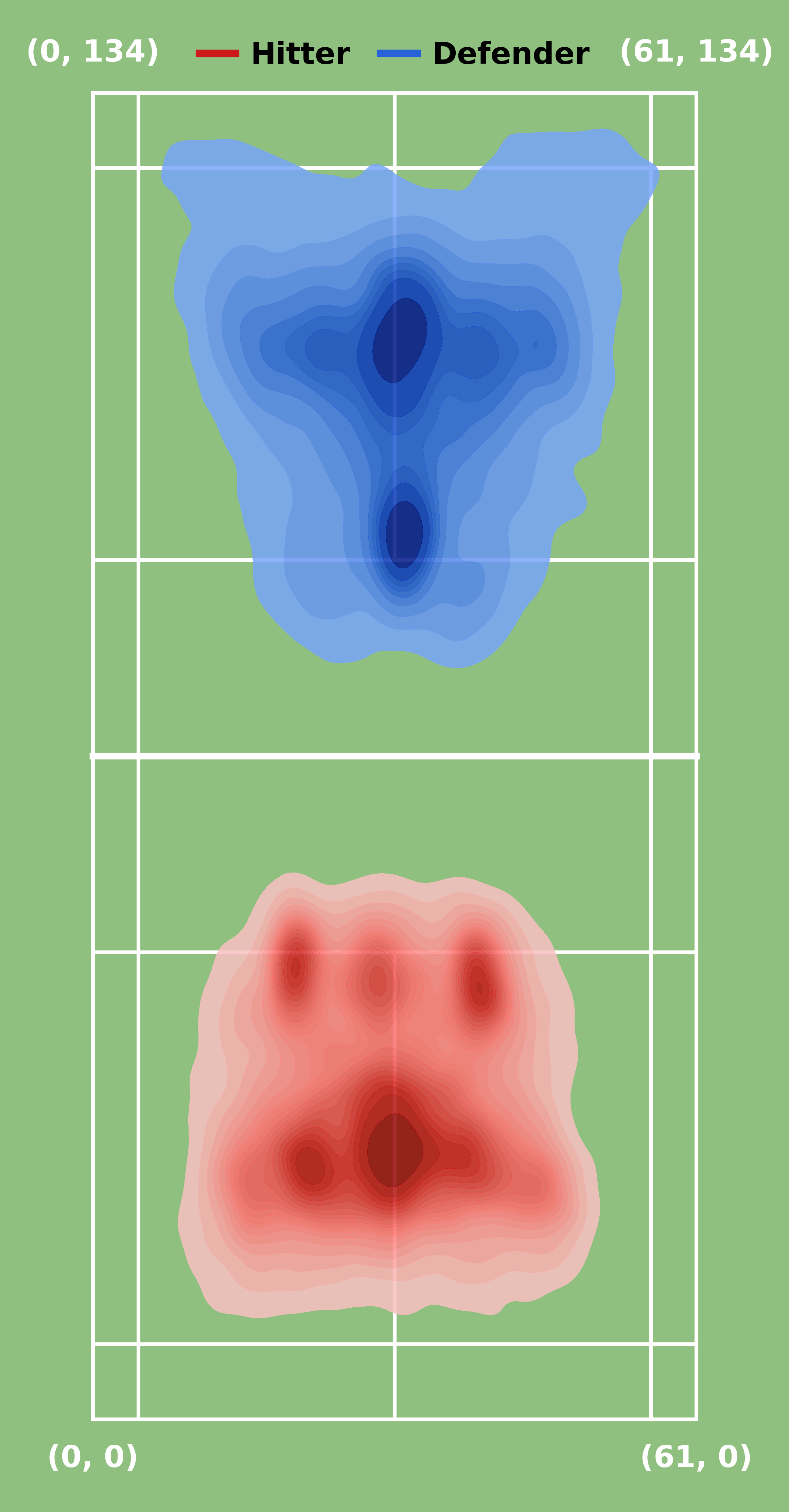}
        \caption{Net Shot position}
        \label{fig:net_pos}
    \end{subfigure}
    
    \caption{Position distribution heatmaps for the remaining five shot types.}
    \label{fig:remaining_heatmaps}
    \vspace{-0.5em}
\end{figure*}

\vspace{-0.5em}
 To further investigate the relationship between tactical intent and court coverage, Figure~\ref{fig:remaining_heatmaps} presents the position distribution heatmaps across the remaining shot types. Observing the spatial patterns, these shot types can be categorized into groups with similar collaborative formations. For instance, the distributions for "Drop" and "Clear" exhibit a offensive positioning seen in the "Smash". "Drive" exhibits a similar defensive pattern to the "Lift", while "Push" and "Net Shot" form another distinct spatial cluster.

\begin{figure*}[htbp]
    \centering
    \begin{subfigure}{0.22\textwidth}
        \includegraphics[width=\linewidth]{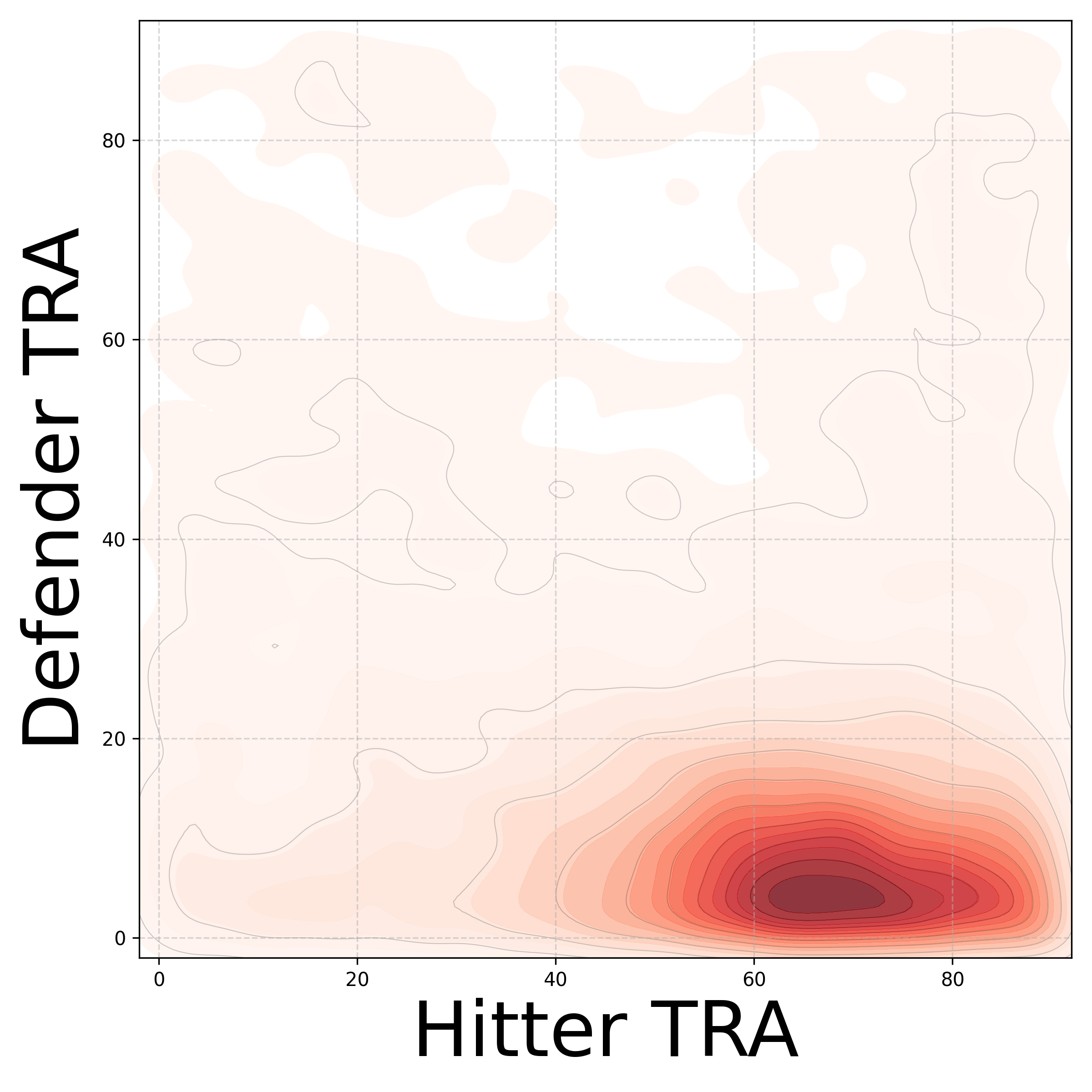}
        \caption{Smash}
        \label{fig:smash}
    \end{subfigure}\hfill
    \begin{subfigure}{0.22\textwidth}
        \includegraphics[width=\linewidth]{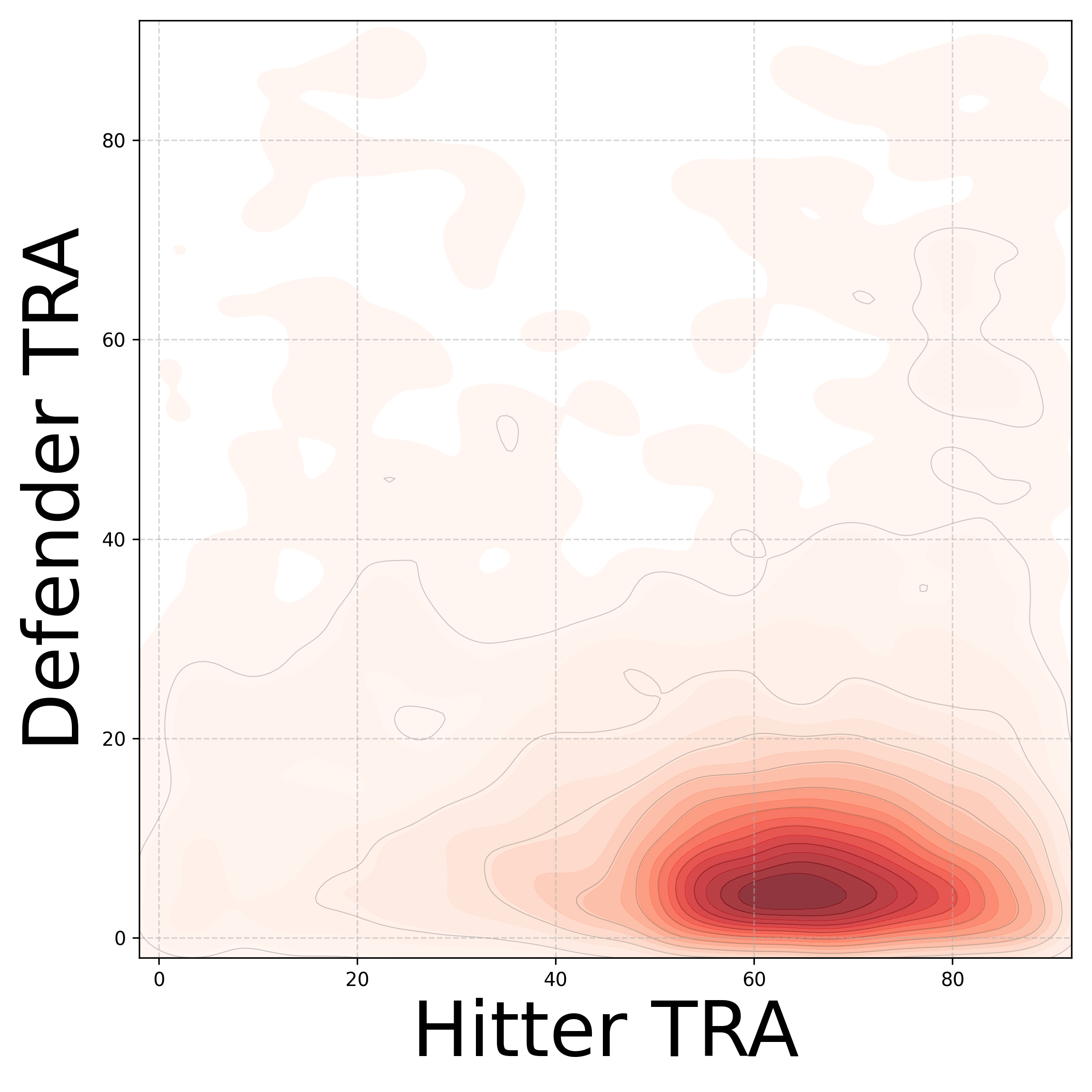}
        \caption{Drop}
        \label{fig:Drop}
    \end{subfigure}\hfill
    \begin{subfigure}{0.22\textwidth}
        \includegraphics[width=\linewidth]{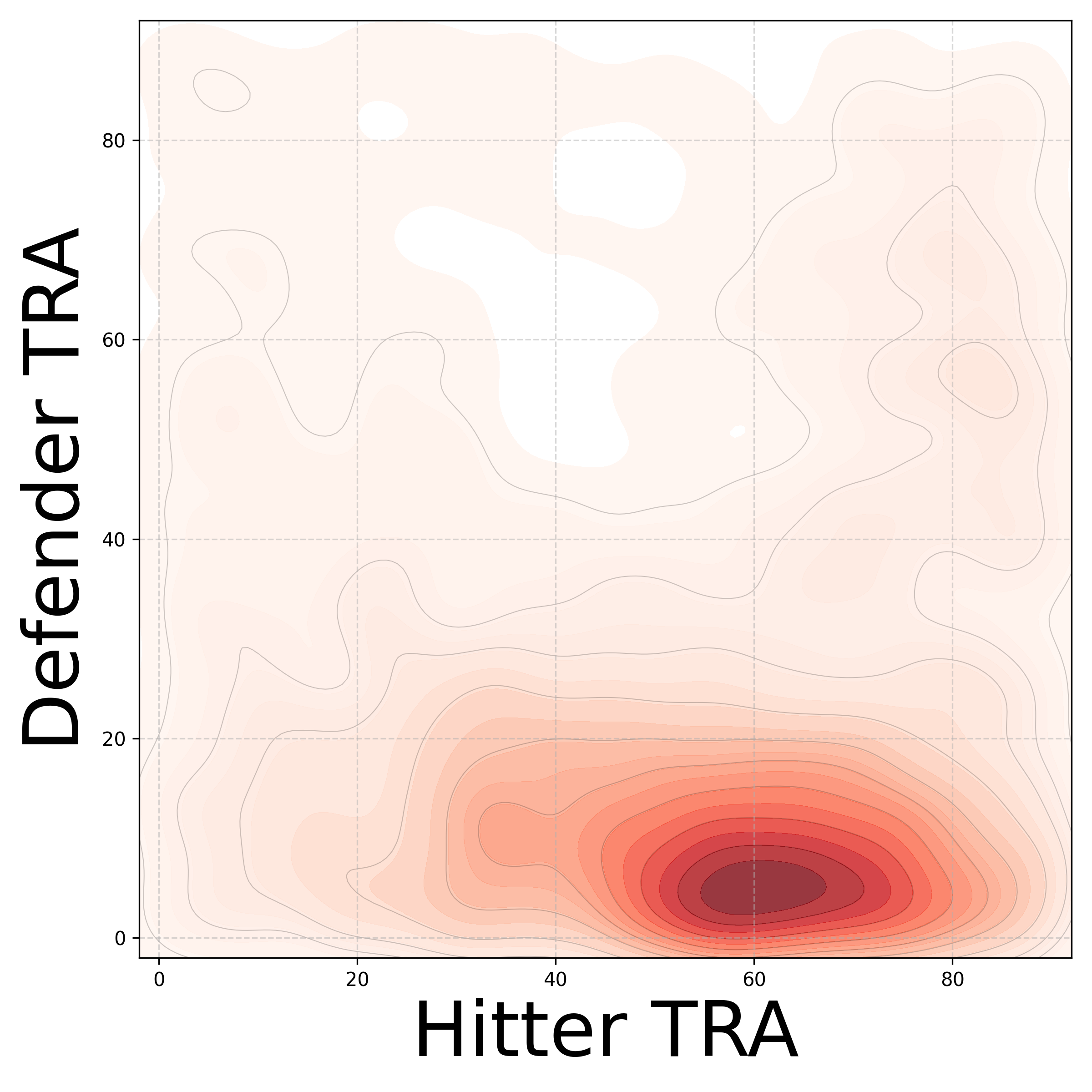}
        \caption{Clear}
        \label{fig:clear}
    \end{subfigure}
    
    \vspace{0.4cm} 
    
    \begin{subfigure}{0.22\textwidth}
        \includegraphics[width=\linewidth]{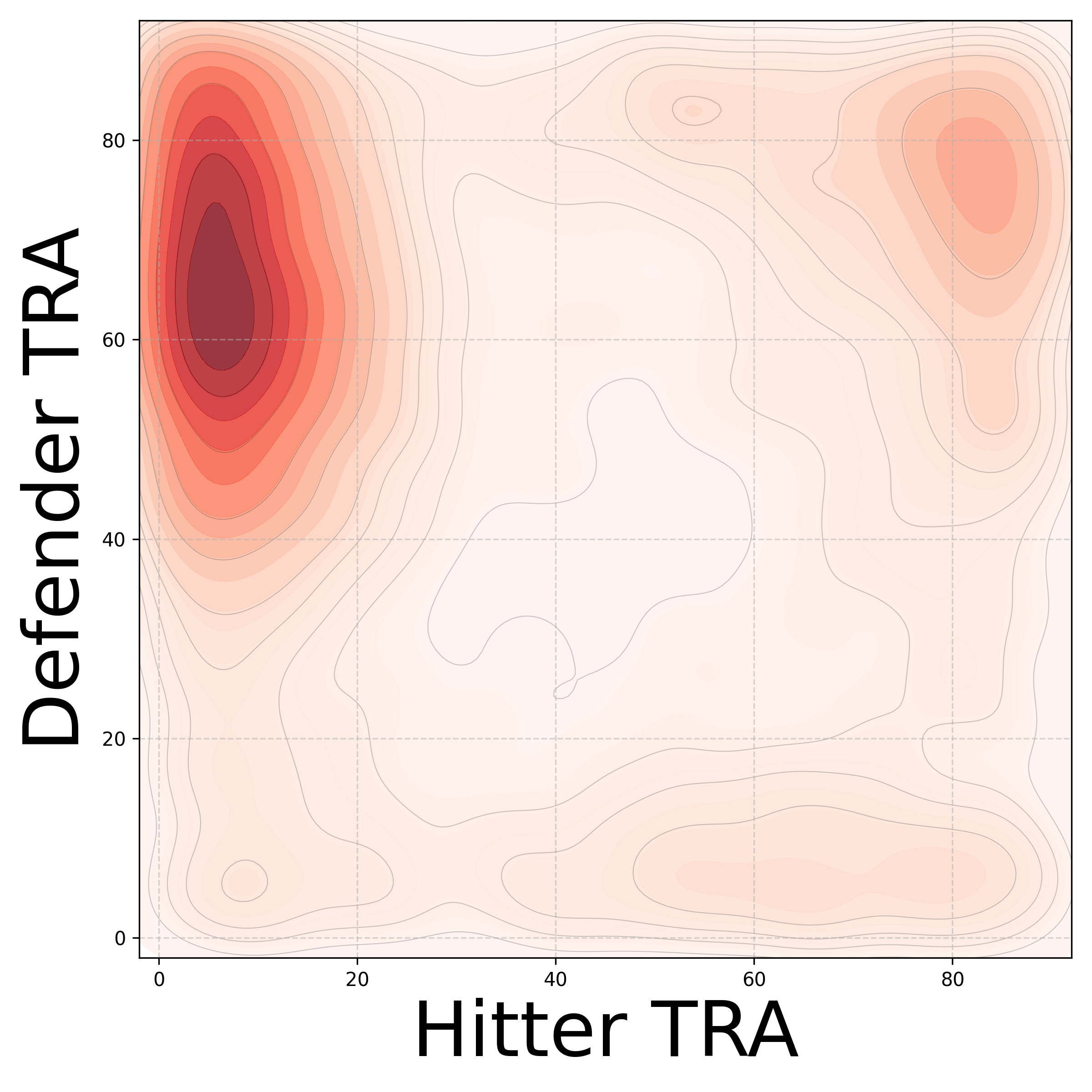}
        \caption{Drive}
        \label{fig:drive}
    \end{subfigure}
    \begin{subfigure}{0.22\textwidth}
        \includegraphics[width=\linewidth]{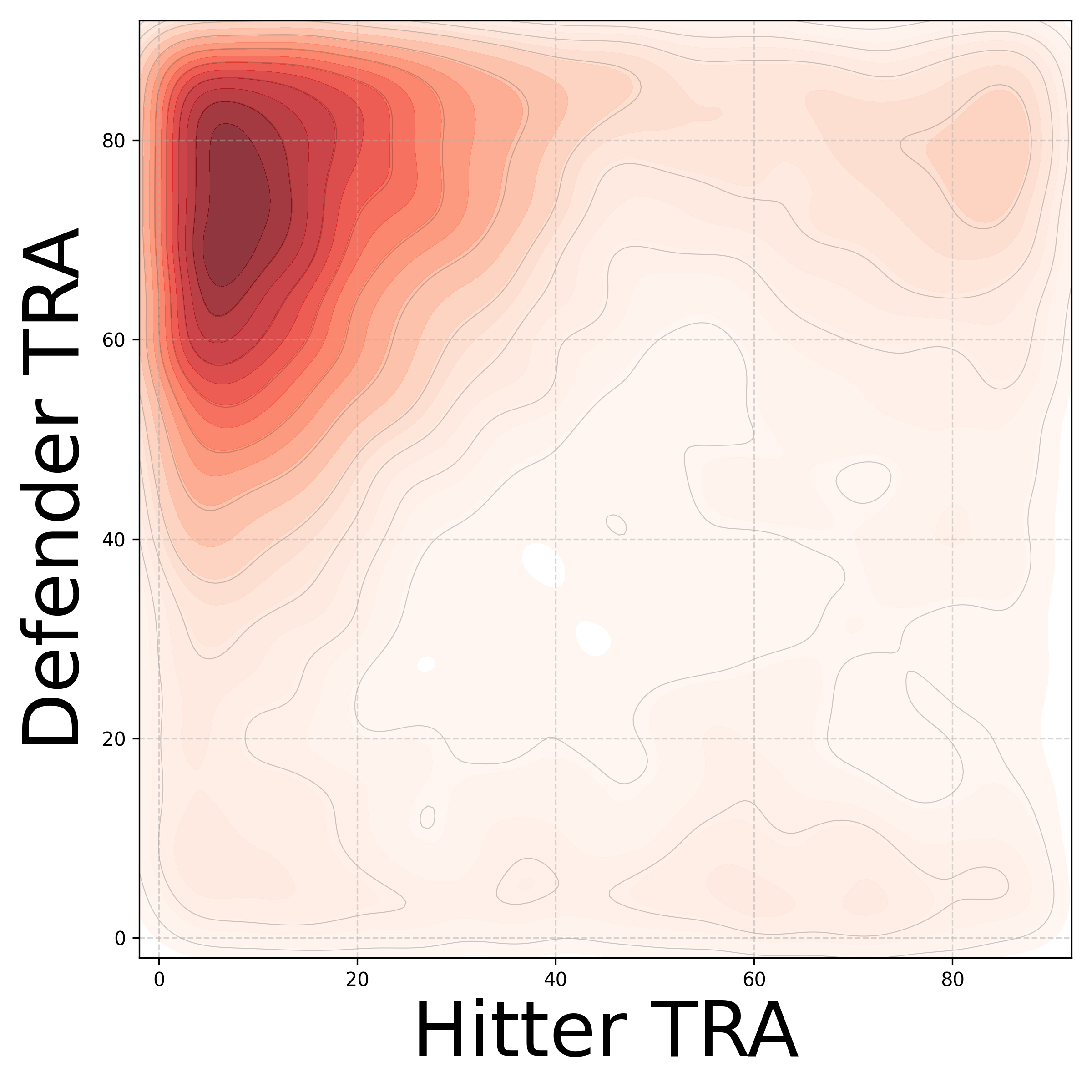}
        \caption{Lift}
        \label{fig:lift}
    \end{subfigure}\hfill
    \begin{subfigure}{0.22\textwidth}
        \includegraphics[width=\linewidth]{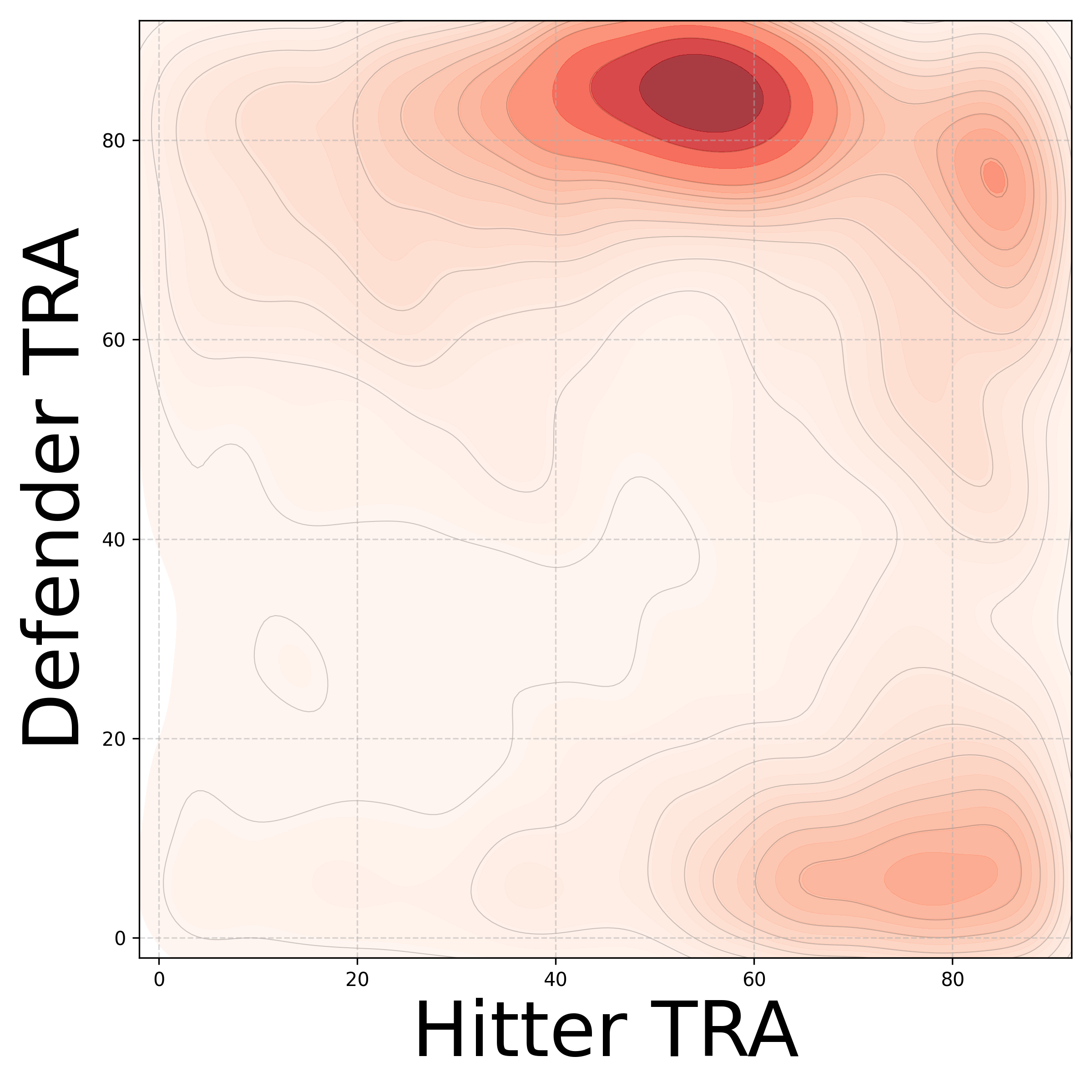}
        \caption{Push}
        \label{fig:push}
    \end{subfigure}
    \hspace{0.05\textwidth} 
    \begin{subfigure}{0.22\textwidth}
        \includegraphics[width=\linewidth]{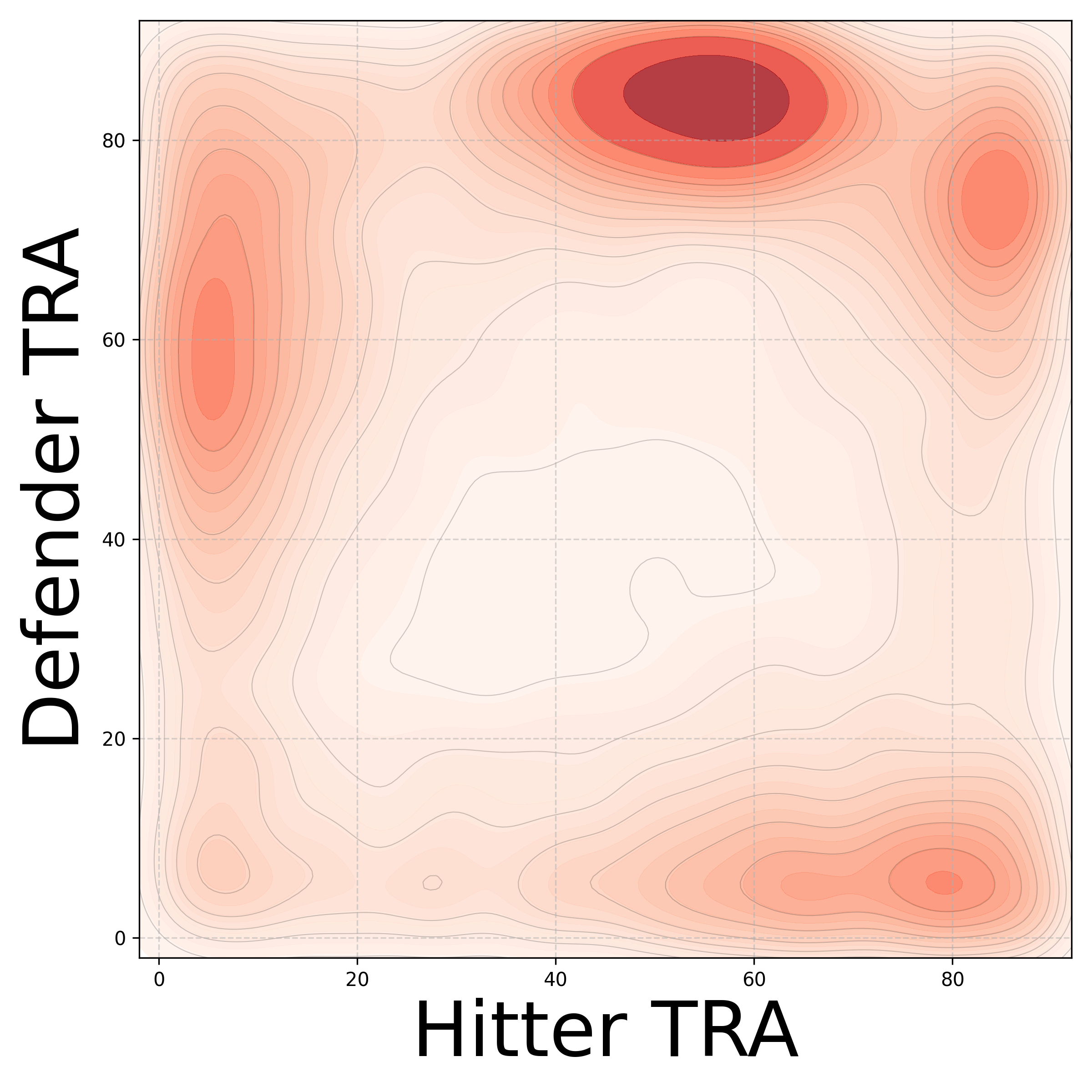}
        \caption{Net Shot}
        \label{fig:net}
    \end{subfigure}
    
    \caption{Angle heatmaps for all shot types.}
    \label{fig:angle heatmap}
    \vspace{-0.5em}
\end{figure*}

Furthermore, to provide a clearer visualization of these team dynamics, the positional data of both offensive and defensive formations is transformed into an angle-based representation, as illustrated in Figure~\ref{fig:angle heatmap}. This angular highlights the tactical phases: shots such as the "Smash", "Drop", and "Clear" are executed at high angles, characterizing a offensive formation. In the second category, defensive shots like the "Lift" and "Drive" are clustered closely around the zero-degree angle, indicating a defensive formation. Finally, instances where both teams simultaneously occupy high angles represent transitional phases where formations are actively shifting.
\subsection{Model Prediction with Angle}
\vspace{-0.5em}
\begin{figure}[H]  
  \centering
  \includegraphics[width=0.4\textwidth]{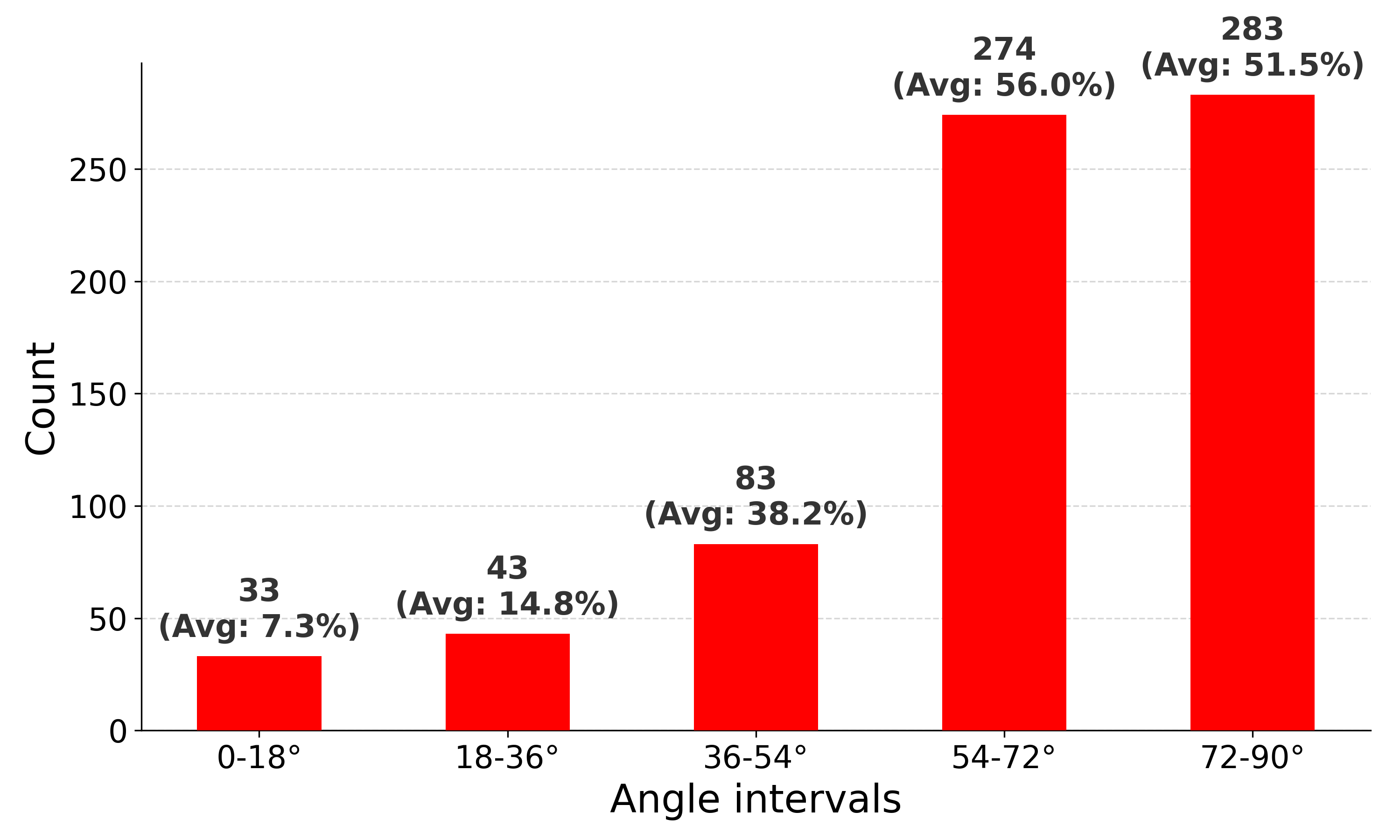}
  \caption{Smash prediction with the angle (threshold=0.8).}
  \label{fig:smash_prediction}
\end{figure}
\subsection{Heatmap}
\begin{figure}[H]  
  \centering
  \includegraphics[width=0.3\textwidth]{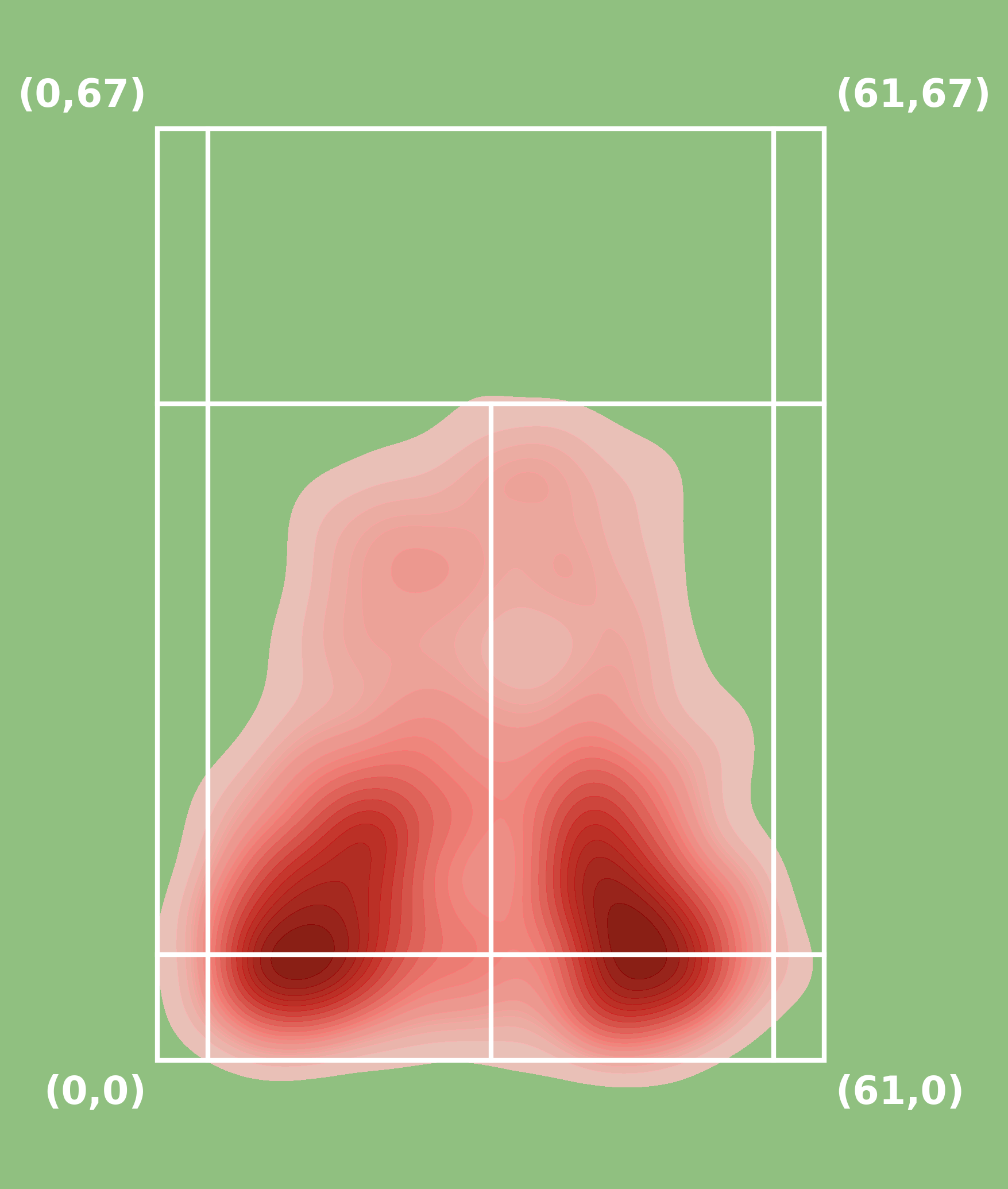}
  \caption{Heatmap with the hitter's smash probability.}
  \label{fig:heatmap}
\end{figure}

\vspace{-0.5em}
To visualize the spatial distribution of high-probability smashes, Figure \ref{fig:heatmap} presents a heatmap generated by weighting the hitter's location with the predicted probability.
The heatmap reveals two distinct dark red zones in the left and right sectors of the backcourt, indicating that high-probability smash opportunities are heavily concentrated in these areas. In contrast to the midcourt, the smash distribution prefers the backcourt, a finding that is consistent with the angular analysis discussed previously.

\subsection{High Probability and Low Probability Position}
In addition to the hitter's location heatmap, a spatial distribution plot analyzing teammate positioning relative to smash probability is introduced.

To enhance visualization, probabilities exceeding 0.8 are represented by red lines, while probabilities below 0.005 are depicted by blue lines.

\begin{figure}[H] 
  \centering
  \begin{subfigure}[b]{0.23\textwidth}
    \centering
    \includegraphics[width=\textwidth]{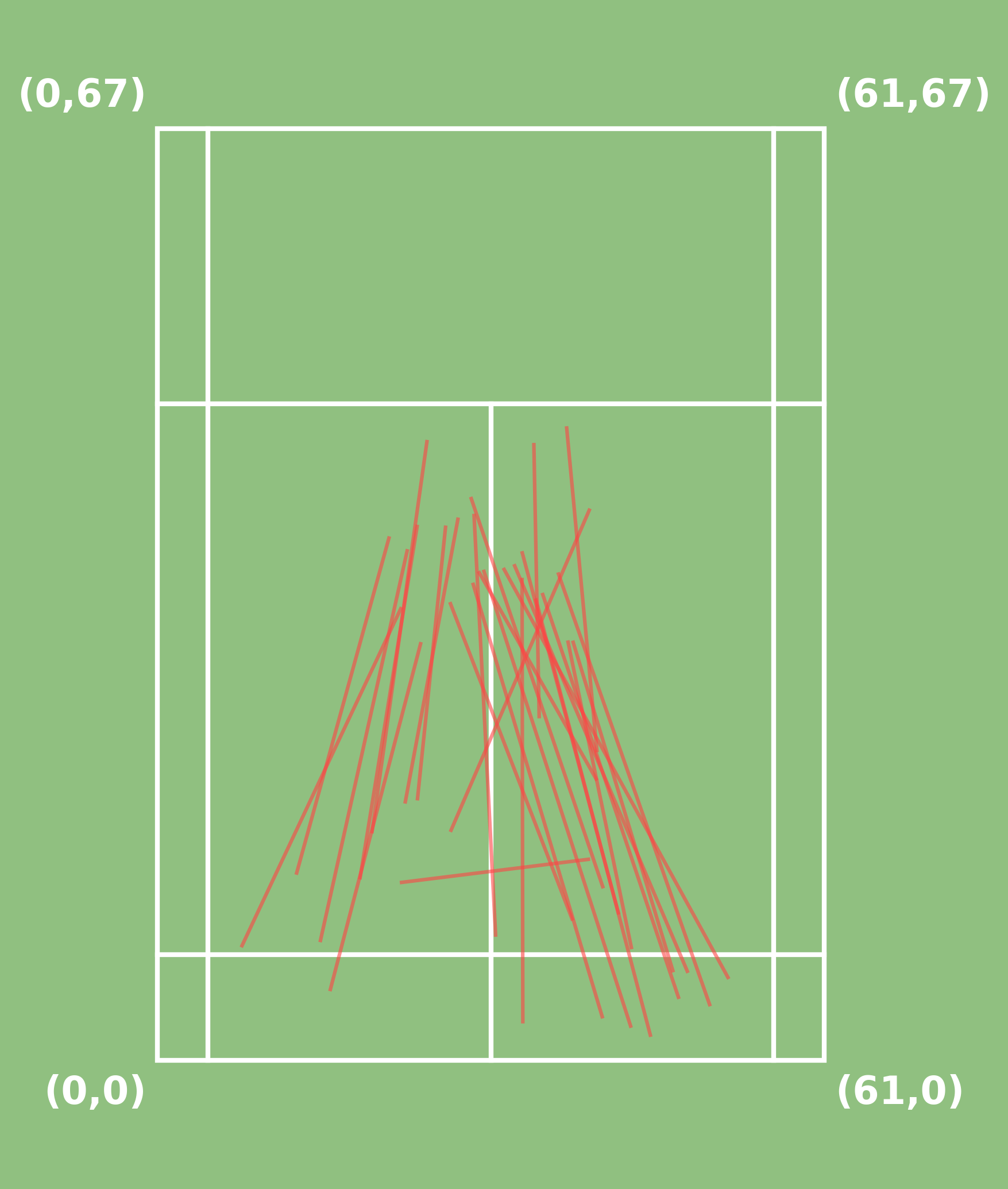}
    \caption{High probability formation}
    \label{fig:teammates_red}
  \end{subfigure}
  \hspace{1.5cm}
  \begin{subfigure}[b]{0.23\textwidth}
    \centering
    \includegraphics[width=\textwidth]{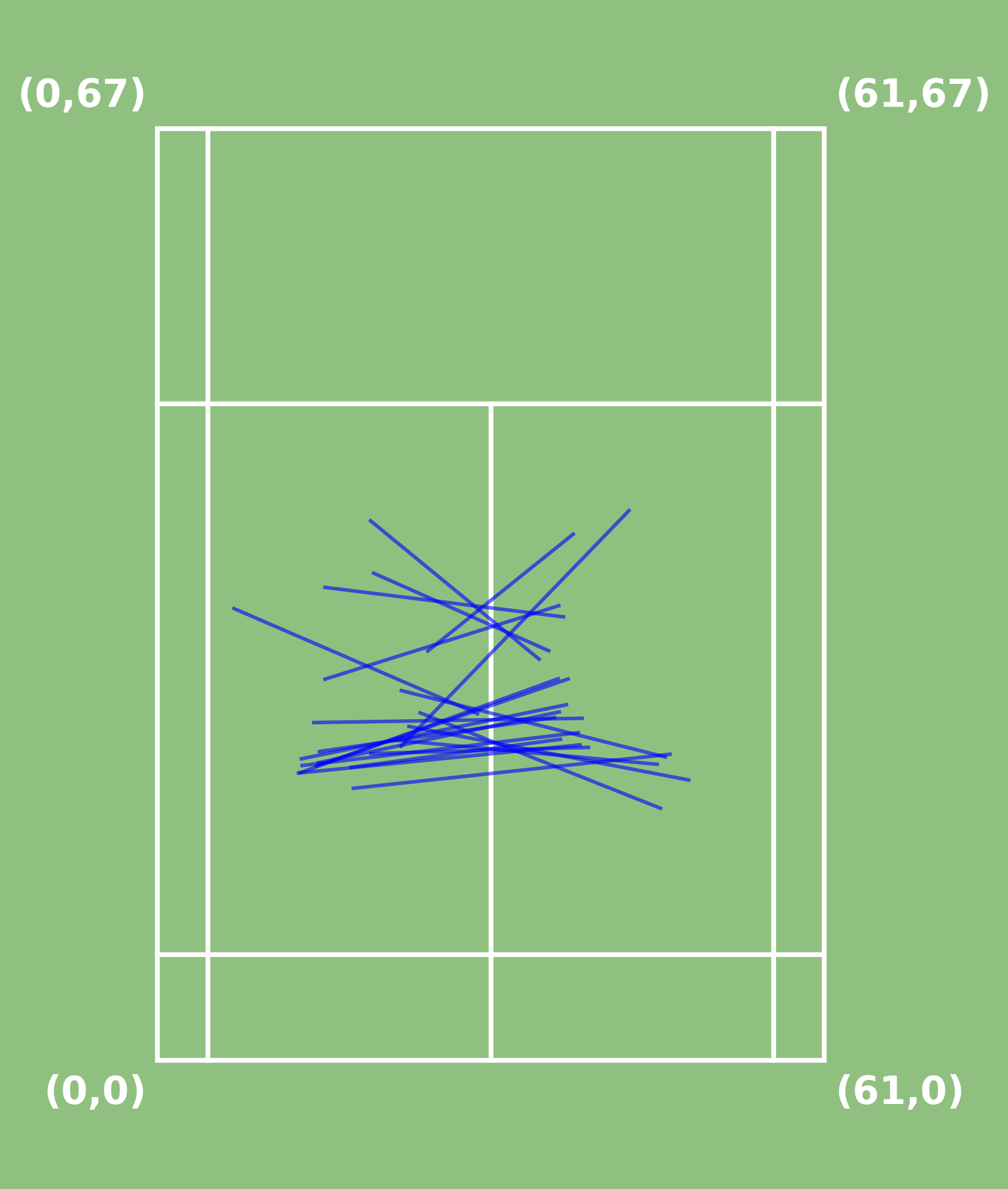}
    \caption{Low probability formation}
    \label{fig:teammates_blue}
  \end{subfigure}

  \caption{Comparison of teammate formations.}
  \label{fig:teammates_comparison}
\end{figure}

As illustrated in Figure \ref{fig:teammates_red} and \ref{fig:teammates_blue} , a distinct pattern emerges: the blue lines correspond to a side-by-side formation, whereas the red lines represent an offensive formation. This observation aligns with previous findings, where the blue zones are concentrated near the mid-court, and the red zones are predominantly located in the backcourt.

\clearpage
\begin{figure}[H]
  \centering
  \includegraphics[width=0.8\textwidth]{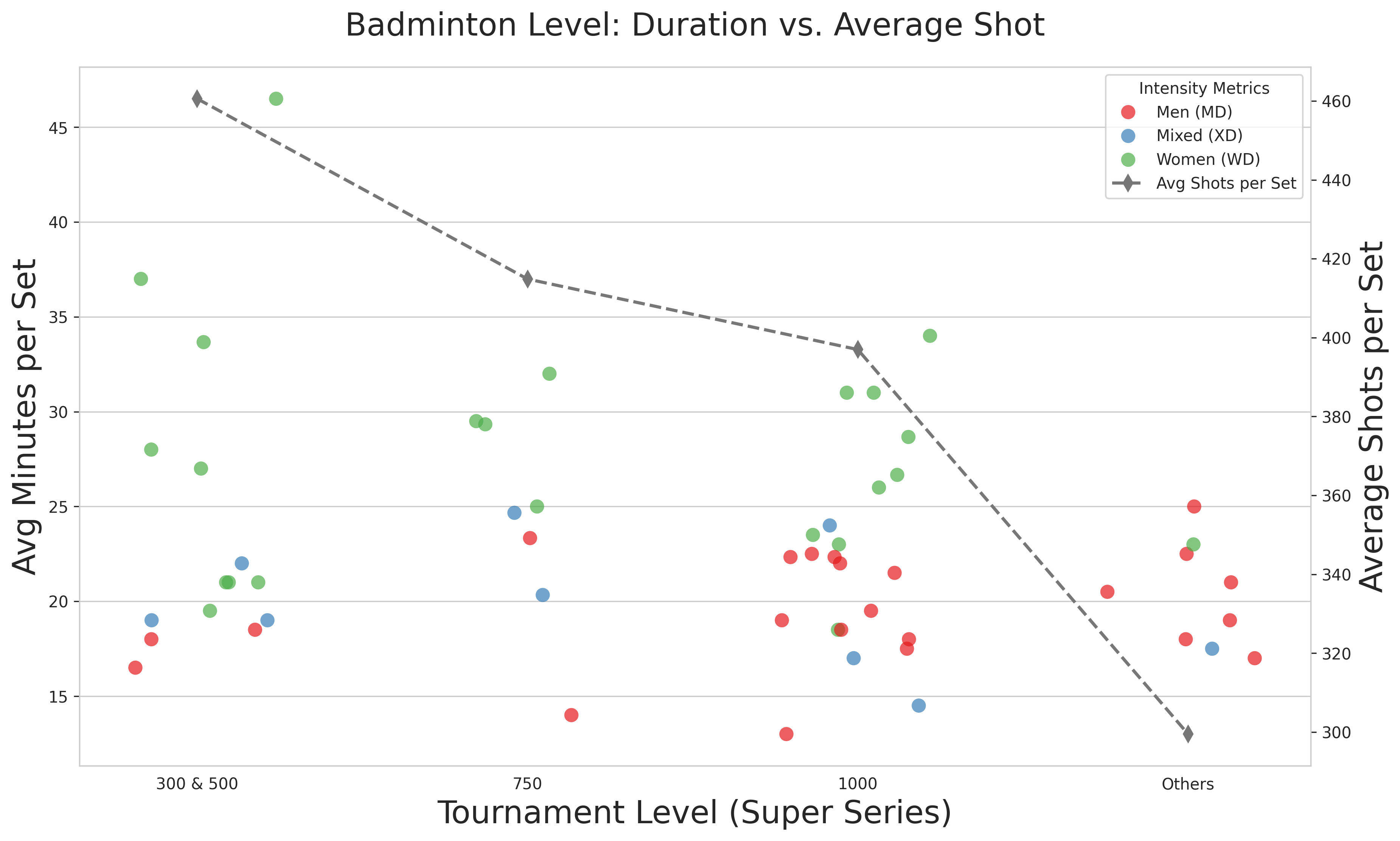}
  \caption{Matches analysis.}
  \label{fig:match_analysis}
\end{figure}

\subsection{Compare Ground Truth vs Each Prediction}
\begin{figure}[H]
  \centering
  \includegraphics[width=0.5\textwidth]{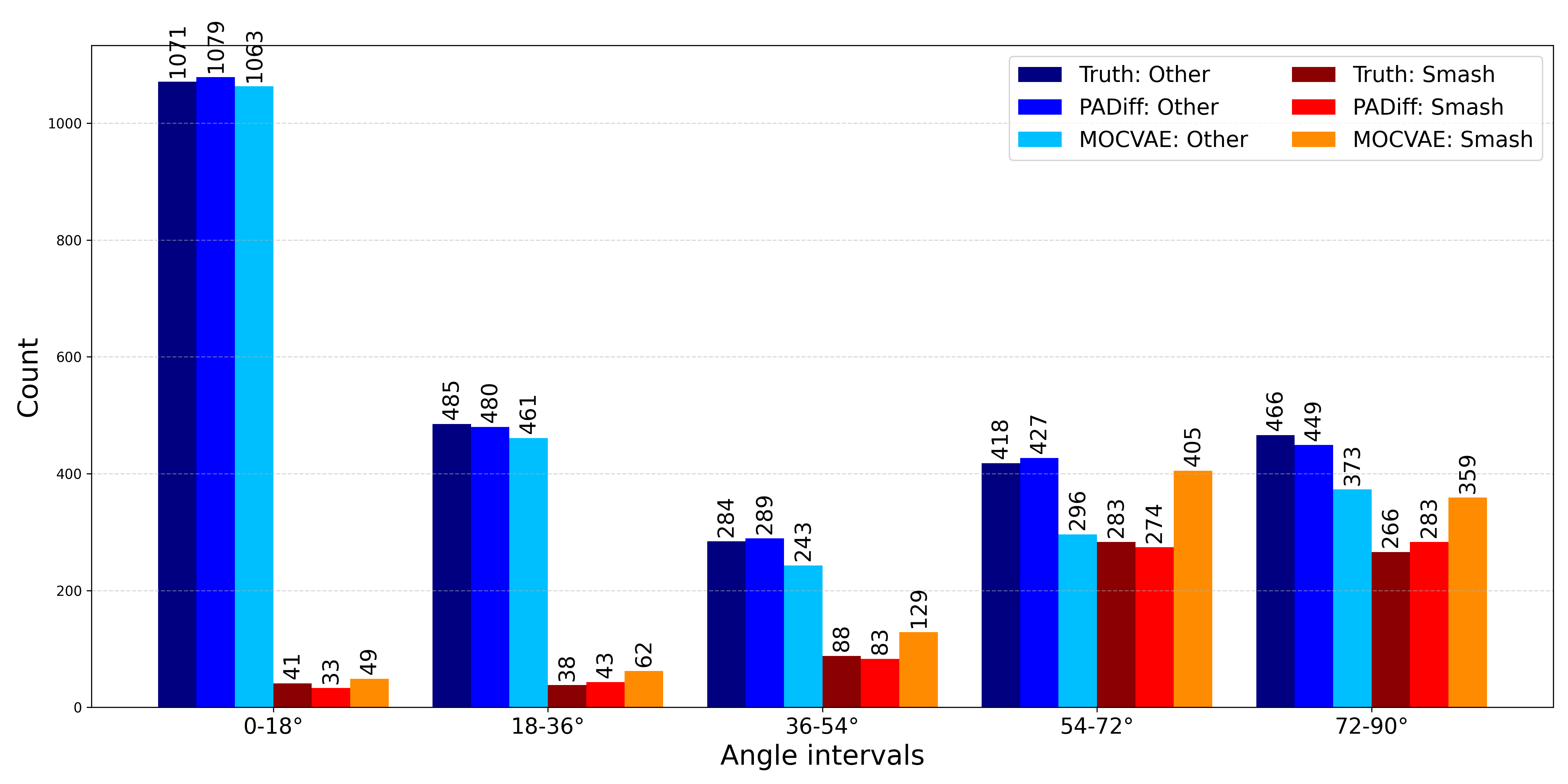}
  \caption{Ground truth v.s. prediction.}
  \label{fig:truth_pred_comparison}
\end{figure}

\vspace{-0.5em}
To visualize the results, the predictions are split by angular intervals. Figure \ref{fig:truth_pred_comparison} presents a comparison among the ground truth, the angle-based model, which is PADiff, and MoCVAE, which serves as the doubles-specific baseline in this study. In the figure, the darkest bars represent the ground truth. The medium-shaded bars correspond to the prediction of the angle-based model, using a probability threshold of 0.8. The lightest bars indicate the predictions from MoCVAE, where a shot is classified as a smash if it yields the highest probability among all shot types. In each interval, the predicted distribution of the angle-based model aligns with the ground truth. This consistency not only validates the model's accuracy but also shows the importance of angular features in predicting the smash shot type.
\begin{table}[H]
    \centering
    \caption{Comparison of smash or non-smash.}
    \label{tab:smash_counting_comparison}
    \begin{tabular}{lccccccc} 
        \toprule
        \multirow{2}{*}{\textbf{Angle}}
        & \multicolumn{3}{c}{\textbf{Smash}} & \multicolumn{3}{c}{\textbf{Non-smash}}\\
        \cmidrule(lr){2-4} 
        \cmidrule(lr){5-7}
        &\textbf{GT} & \textbf{PAD} &\textbf{MoC}
        &\textbf{GT} & \textbf{PAD} &\textbf{MoC} \\
        \midrule
        0-18°      & 41 & 33 & 49 & 1071& 1079& 1063 \\
        18-36°     & 38 & 43 & 62 & 485& 480& 461\\
        36-54°    & 88 & 83 & 129 & 284& 289& 243\\
        54-72°    & 283 & 274 & 405& 418& 427& 296 \\
        72-90°    & 266 & 283 & 359& 466& 449& 373 \\
        \bottomrule
    \end{tabular}
\end{table}




\subsection{Spatial Discretization and Aerodynamic Weighting for \metrixDefcoordinationshort}
\label{app:defensive_coordination}
\vspace{-0.5em}
\textbf{Court Discretization and Voronoi Partitioning:} To compute the \metrixDefcoordinationshort, we discretize the defensive half-court into a $61 \times 67$ grid, with each cell $g$ representing a $10\text{ cm} \times 10\text{ cm}$ area. Using the 2D tracking coordinates, each cell is assigned to the nearest player based on Euclidean distance. This nearest-neighbor assignment approximates the Voronoi responsibility zones ($V_1$ and $V_2$) applied in the threat load summation.

\textbf{Physical Interpretation of the Aerodynamic Constant:} Regarding the dynamic threat weight function $W(g)$, the exponential decay models the non-linear aerodynamic deceleration of shuttlecocks \cite{shuttlecockdecay}. The constant $3.35$ acts as a kinematic half-distance decay factor. Physically, this indicates that the threat level, associated with residual velocity and reduced defender reaction time, decreases by half for every $3.35\text{ meters}$ the shuttlecock travels from the attack origin ($S$). Consequently, this formulation assigns higher defensive weights to regions closer to the attack origin, reflecting the uneven spatial pressure experienced by the defending team.

\section{Detailed Match Statistics} \label{app:match_stats}

Please refer to Table \ref{tab:match_info_1} and Table \ref{tab:match_info_2} for the comprehensive information of each match.
\begin{sidewaystable}
    \centering
    \caption{Detailed information of each labeled match (Part I).}
    \label{tab:match_info_1}
    \begingroup
    \setlength{\tabcolsep}{2.5pt}
    \resizebox{\textheight}{!}{%
    \begin{tabular}{l l c c c l l}
        \toprule
        Tournament & Round & Year & Set & Duration (m) & Winner & Loser \\
        \midrule
        GWANGJU YONEX Korea Masters 2022 & Finals & 2022 & 2 & 37 & Kim Gi Jung \& Kim Sa Rang & Liu Yu Chen \& Ou Xuan Yi \\
        GWANGJU YONEX Korea Masters 2022 & Finals & 2022 & 2 & 63 & Kim So Yeong \& Kong Hee Yong & Baek Ha Na \& Lee Yu Lim \\
        DAIHATSU Indonesia Masters 2022 & Finals & 2022 & 2 & 42 & Fajar Alfian \& Muhammad Rian Ardianto & Liang Wei Keng \& Wang Chang \\
        DAIHATSU Indonesia Masters 2022 & Semi-finals & 2022 & 2 & 33 & Liang Wei Keng \& Wang Chang & Kevin Sanjaya Sukamuljo \& Marcus Fernaldi Gideon \\
        DAIHATSU Indonesia Masters 2022 & Finals & 2022 & 2 & 38 & Zheng Si Wei \& Huang Ya Qiong & Thom Gicquel \& Delphine Delrue \\
        Asia Championships 2023 & Round-of-32 & 2023 & 2 & 37 & Fukushima Yuki \& Hirota Sayaka & Lee Chia Hsin \& Teng Chun Hsun \\
        Asia Championships 2023 & Round-of-32 & 2023 & 3 & 51 & Goh Soon Huat \& Lai Shevon Jemie & Ye Hong Wei \& Lee Chia Hsin \\
        YONEX All England Open 2023 & Semi-finals & 2023 & 2 & 46 & Baek Ha Na \& Lee So Hee & Treesa Jolly \& Gayatri Gopichand \\
        YONEX All England Open 2023 & Quarter-finals & 2023 & 3 & 93 & Baek Ha Na \& Lee So Hee & Apriyani Rahayu \& Siti Fadia Silva Ramadhanti \\
        YONEX All England Open 2023 & Quarter-finals & 2023 & 3 & 93 & Kim So Yeong \& Kong Hee Yong & Chen Qing Chen \& Jia Yi Fan \\
        YONEX German 2023 & Semi-finals & 2023 & 2 & 52 & Matsuyama Nami \& Shida Chiharu & Pearly Tan \& Thinaah Muralitharan \\
        YONEX German 2023 & Finals & 2023 & 2 & 72 & Baek Ha Na \& Lee So Hee & Matsuyama Nami \& Shida Chiharu \\
        PERODUA Malaysia Masters 2023 & Finals & 2023 & 3 & 101 & Baek Ha Na \& Lee So Hee & Pearly Tan \& Thinaah Muralitharan \\
        PERODUA Malaysia Masters 2023 & Quarter-finals & 2023 & 3 & 125 & Pearly Tan \& Thinaah Muralitharan & Fukushima Yuki \& Hirota Sayaka \\
        PERODUA Malaysia Masters 2023 & Semi-finals & 2023 & 2 & 105 & Pearly Tan \& Thinaah Muralitharan & Jeong Na Eun \& Kim Hye Jeong \\
        Thailand Open 2023 & Semi-finals & 2023 & 2 & 43 & Kim So Yeong \& Kong Hee Yong & Jongkolphan Kititharakul \& Rawinda Prajongjai \\
        Thailand Open 2023 & Finals & 2023 & 2 & 38 & Kim So Yeong \& Kong Hee Yong & Benyapa Aimsaard \& Nuntakarn Aimsaard \\
        Thailand Open 2023 & Semi-finals & 2023 & 2 & 34 & Benyapa Aimsaard \& Nuntakarn Aimsaard & Baek Ha Na \& Lee So Hee \\
        VICTOR China Open 2023 & Semi-finals & 2023 & 3 & 77 & Liang Wei Keng \& Wang Chang & Takuro Hoki \& Yugo Kobayashi \\
        VICTOR China Open 2023 & Quarter-finals & 2023 & 2 & 48 & Liang Wei Keng \& Wang Chang & Liu Yu Chen \& Ou Xuan Yi \\
        VICTOR China Open 2023 & Finals & 2023 & 2 & 26 & Liang Wei Keng \& Wang Chang & Aaron Chia \& Soh Wooi Yik \\
        KAPAL API GROUP Indonesia Open 2023 & Finals & 2023 & 2 & 61 & Satwiksairaj Rankireddy \& Chirag Shetty & Aaron Chia \& Soh Wooi Yik \\
        KAPAL API GROUP Indonesia Open 2023 & Semi-finals & 2023 & 3 & 80 & Aaron Chia \& Soh Wooi Yik & Pramudya Kusumawardana \& Yeremia Erich Yoche Yacob Rambitan \\
        KAPAL API GROUP Indonesia Open 2023 & Semi-finals & 2023 & 3 & 52 & Satwiksairaj Rankireddy \& Chirag Shetty & Kang Min Hyuk \& Seo Seung Jae \\
        KAPAL API GROUP Indonesia Open 2023 & Semi-finals & 2023 & 3 & 89 & Fukushima Yuki \& Hirota Sayaka & Matsumoto Mayu \& Nagahara Wakana \\
        KAPAL API GROUP Indonesia Open 2023 & Semi-finals & 2023 & 2 & 60 & Baek Ha Na \& Lee So Hee & Jeong Na Eun \& Kim Hye Jeong \\
        KAPAL API GROUP Indonesia Open 2023 & Finals & 2023 & 2 & 64 & Baek Ha Na \& Lee So Hee & Fukushima Yuki \& Hirota Sayaka \\
        DAIHATSU Japan Open 2023 & Quarter-finals & 2023 & 3 & 80 & Lee Yang \& Wang Chi Lin & Satwiksairaj Rankireddy \& Chirag Shetty \\
        KAPAL API GROUP Indonesia Open 2023 & Round-of-32 & 2023 & 3 & 76 & Koga Akira \& Saito Taichi & Lee Yang \& Wang Chi Lin \\
        PETRONAS Malaysia Open 2024 & Semi-finals & 2024 & 2 & 58 & Liu Sheng Shu \& Tan Ning & Iwanaga Rin \& Nakanishi Kie \\
        YONEX French Open 2024 & Semi-finals & 2024 & 2 & 65 & Chen Qing Chen \& Jia Yi Fan & Matsumoto Mayu \& Nagahara Wakana \\
        YONEX French Open 2024 & Finals & 2024 & 3 & 107 & Chen Qing Chen \& Jia Yi Fan & Matsuyama Nami \& Shida Chiharu \\
        YONEX French Open 2024 & Round-of-32 & 2024 & 3 & 53 & Liang Wei Keng \& Wang Chang & Lee Yang \& Wang Chi Lin \\
        YONEX SUNRISE India Open 2024 & Semi-finals & 2024 & 3 & 84 & Dechapol Puavaranukroh \& Sapsiree Taerattanachai & Ye Hong Wei \& Lee Chia Hsin \\
        YONEX Swiss Open 2024 & Semi-finals & 2024 & 2 & 48 & Chen Tang Jie \& Toh Ee Wei & Ye Hong Wei \& Lee Chia Hsin \\
        YONEX All England Open 2024 & Round-of-32 & 2024 & 2 & 38 & Ye Hong Wei \& Lee Chia Hsin & Seo Seung Jae \& Chae Yu Jung \\
        YONEX All England Open 2024 & Quarter-finals & 2024 & 2 & 61 & Fajar Alfian \& Muhammad Rian Ardianto & Lee Yang \& Wang Chi Lin \\
        VICTOR Denmark Open 2023 & Quarter-finals & 2023 & 3 & 71 & Mathias Christiansen \& Alexandra Bøje & Ye Hong Wei \& Lee Chia Hsin \\
        YONEX All England Open 2024 & Semi-finals & 2024 & 2 & 47 & Aaron Chia \& Soh Wooi Yik & Lee Jhe Huei \& Yang Po Hsuan \\
        YONEX All England Open 2024 & Semi-finals & 2024 & 3 & 95 & Baek Ha Na \& Lee So Hee & Kim So Yeong \& Kong Hee Yong \\
        \bottomrule
    \end{tabular}%
    }
    \endgroup
\end{sidewaystable}

\begin{sidewaystable}
    \centering
    \caption{Detailed information of each labeled match (Part II).}
    \label{tab:match_info_2}
    \begingroup
    \setlength{\tabcolsep}{2.5pt}
    \resizebox{\textheight}{!}{%
    \begin{tabular}{l l c c c l l}
        \toprule
        Tournament & Round & Year & Set & Duration (m) & Winner & Loser \\
        \midrule
        Paris Olympic 2024 & Group-Stage & 2024 & 2 & 36 & Tang Chun Man \& Tse Ying Suet & Ye Hong Wei \& Lee Chia Hsin \\
        Paris Olympic 2024 & Group-Stage & 2024 & 2 & 33 & Lee Yang \& Wang Chi Lin & Hoki Takuro \& Kobayashi Yugo \\
        Paris Olympic 2024 & Group-Stage & 2024 & 2 & 46 & Satwiksairaj Rankireddy \& Chirag Shetty & Lucas Corvee \& Ronan Labar \\
        Paris Olympic 2024 & Group-Stage & 2024 & 2 & 43 & Astrup Kim \& Rasmussen Anders Skaarup & Chiu Vinson \& Yuan Joshua \\
        Paris Olympic 2024 & Group-Stage & 2024 & 2 & 47 & Liang Wei Keng \& Wang Chang & Aaron Chia \& Soh Wooi Yik \\
        Paris Olympic 2024 & Group-Stage & 2024 & 2 & 39 & Satwiksairaj Rankireddy \& Chirag Shetty & Fajar Alfian \& Muhammad Rian Ardianto \\
        Paris Olympic 2024 & Quarter-finals & 2024 & 2 & 51 & Liang Wei Keng \& Wang Chang & Fajar Alfian \& Muhammad Rian Ardianto \\
        Paris Olympic 2024 & Quarter-finals & 2024 & 2 & 51 & Liang Wei Keng \& Wang Chang & Fajar Alfian \& Muhammad Rian Ardianto \\
        Paris Olympic Games & Group-Stage & 2024 & 2 & 38 & Satwiksairaj Rankireddy \& Chirag Shetty & Fajar Alfian \& Muhammad Rian Ardianto \\
        Paris Olympic Games & Quarter-finals & 2024 & 2 & 50 & Liang Wei Keng \& Wang Chang & Fajar Alfian \& Muhammad Rian Ardianto \\
        DAIHATSU Japan Open 2024 & Finals & 2024 & 2 & 64 & Liu Sheng Shu \& Tan Ning & Baek Ha Na \& Lee So Hee \\
        DAIHATSU Japan Open 2024 & Semi-finals & 2024 & 2 & 59 & Baek Ha Na \& Lee So Hee & Iwanaga Rin \& Nakanishi Kie \\
        BWF World Tour Finals 2024 & Semi-finals & 2024 & 2 & 68 & Matsuyama Nami \& Shida Chiharu & Jia Yi Fan \& Chen Qing Chen \\
        BWF World Tour Finals 2024 & Semi-finals & 2024 & 2 & 39 & Zheng Si Wei \& Huang Ya Qiong & Jiang Zhen Bang \& Wei Ya Xin \\
        YONEX All England Open 2025 & Semi-finals & 2025 & 2 & 46 & Matsuyama Nami \& Shida Chiharu & Liu Sheng Shu \& Tan Ning \\
        YONEX All England Open 2025 & Finals & 2025 & 2 & 42 & Kim Won Ho \& Seo Seung Jae & Leo Rolly Carnando \& Bagas Maulana \\
        PETRONAS Malaysia Open 2025 & Semi-finals & 2025 & 2 & 61 & Fukushima Yuki \& Matsumoto Mayu & Li Yi Jing \& Luo Xu Min \\
        KFF Singapore Badminton Open 2025 & Semi-finals & 2025 & 2 & 42 & Kim Won Ho \& Seo Seung Jae & Astrup Kim \& Rasmussen Anders Skaarup \\
        KFF Singapore Badminton Open 2025 & Finals & 2025 & 2 & 34 & Dechapol Puavaranukroh \& Supissara Paewsampran & Tang Chun Man \& Tse Ying Suet \\
        VICTOR China Open 2025 & Finals & 2025 & 2 & 35 & Fajar Alfian \& Muhammad Shohibul Fikri & Aaron Chia \& Soh Wooi Yik \\
        VICTOR China Open 2025 & Semi-finals & 2025 & 2 & 37 & Fajar Alfian \& Muhammad Shohibul Fikri & Liang Wei Keng \& Wang Chang \\
        KAPAL API Indonesia Open 2025 & Finals & 2025 & 2 & 48 & Thom Gicquel \& Delphine Delrue & Dechapol Puavaranukroh \& Supissara Paewsampran \\
        \bottomrule
    \end{tabular}%
    }
    \endgroup
\end{sidewaystable}

\section{Annotation Process}
\label{sec:Detail Annotation}
The annotation process adapts the protocols established in ShuttleSet \cite{wang2023shuttleset}, utilizing the shot-by-shot ($S^2$) labeling tool \cite{s2labeling} to extract stroke-level data. The pipeline comprises three main stages, followed by quality control and tactical feature extraction.

\begin{figure}[t]
    \centering
    \includegraphics[width=\linewidth]{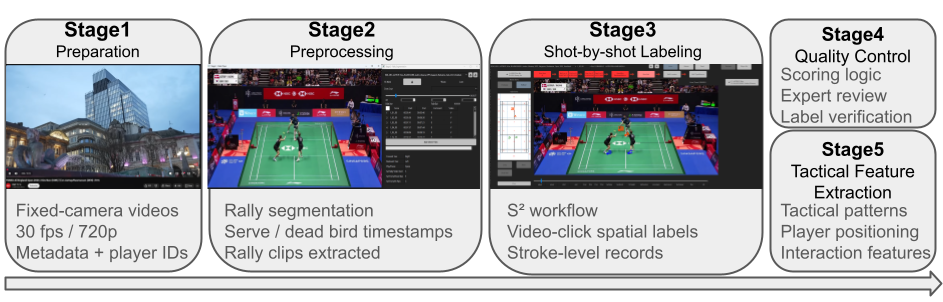}
    \caption{Pipeline of stroke-level annotation and tactical feature extraction.}
    \label{fig:annotation_pipeline}
\end{figure}

\paragraph{Stage 1: Match Video Preparation}
We sourced fixed-camera match videos capturing the four court corners from the public BWF website. To standardize the annotation, all videos were resampled to 30 fps at a resolution of $1280 \times 720$ (720p). Match-level metadata was recorded following the ShuttleSet schema, with additional registration for the identities of the four competing players.

\paragraph{Stage 2: Match Video Preprocessing}
To streamline the pipeline, full-match videos underwent a segmentation phase. Annotators recorded the start (serve) and end (dead bird) timestamps alongside the outcome of each rally. Based on these precise timestamps, the $S^2$ tool extracted individual rally clips. Filtering out inactive periods between rallies effectively reduces downstream annotation complexity.

\paragraph{Stage 3: Shot-by-shot Labeling}
Domain experts annotated the preprocessed clips chronologically using a single-pass workflow. The $S^2$ interface ensures spatial accuracy by allowing annotators to map positions via direct video clicks, generating real-time markers for visual confirmation. For each shot, annotators logged the hitter, timestamp, tactical shot type, stroke characteristics (e.g., forehand/backhand and hitting height), and the $(x, y)$ coordinates of all four players and the shuttle. Iteratively treating the current shot's return time as the start of the next stroke, this continuous cycle captures the dynamic spatial configurations and tactical interactions within a rally.

\paragraph{Quality Control}
To ensure dataset reliability, domain experts randomly sampled and reviewed annotated rallies to verify the accuracy of spatial trajectories, scoring logic, and stroke-level records.

\section{Experimental Setup for Baseline Models}\label{apdx:baseline}
Player Behavior Prediction task is formulated to forecast three primary outputs: future trajectory, angular configuration, and shot type probabilities. 
The prediction task is defined as forecasting future tactical behaviors based on historical context. Specifically, given a sequence of the preceding $n-1$ shots, the objective is to predict the $n$-th shot attributes: trajectory, angle, and shot type probabilities.

The model performance is evaluated using task-specific metrics. For trajectory prediction, \metrixDForcastingfullF \ (\metrixDForcastingshortF) measures the Euclidean distance between the predicted player position and the ground truth at the final timestep. Angular prediction is quantified via \metricDForcastingfullM \ (\metricDForcastingshortM) to assess orientation deviation. Finally, shot type probability prediction is assessed using \metrixDForcastingAcc \ and \metrixDForcastingFs.

For alignment with the task definition, the raw dataset of 54,141 stroke-level records is transformed into 17,440 rally samples using a sliding window approach, with a training-to-testing ratio of 4:1. To establish baseline performance for \DatasetName, two representative models are evaluated: MoCVAE \cite{sung2024mocvae}, and TSCVAE \cite{sung2025player}. 

All experiments were conducted on a machine equipped with an AMD64 Family 25 Model 97 Stepping 2 Authentic AMD 6-Core CPU, an NVIDIA RTX 3060 12GB GPU, and 32GB of RAM.
Given that current baseline models do not explicitly incorporate team dynamics, it is essential to demonstrate the tactical significance of collaborative relationships in doubles badminton. To provide empirical evidence, the spatial distributions of player positions corresponding to distinct tactical intents are analyzed. Specifically, the "Smash" represents an offensive formation, and the "Lift" representing a defensive formation. 

As illustrated in Figure~\ref{fig:smash_pos}, the heatmaps visualize the court coverage of the hitter (indicated in red) and the opposing team (indicated in blue), with color intensity reflecting the frequency of occurrences at specific locations. These visualizations clearly demonstrate that team positioning patterns are highly correlated with the executed shot types, revealing distinct offensive and defensive collaborative formations. This observation reinforces the necessity of integrating multi-player spatial dynamics to better capture tactical intent. Further detailed visualizations and comprehensive analyses are provided in the Appendix \ref{app:spatial_distribution}.


\section{Societal Implications of \DatasetName}
\label{sec:societal_impact}
\DatasetName~introduces the first stroke-level doubles dataset with synchronized four-player coordinates and expert-annotated loss reasons , enabling research directions that were previously infeasible, such as formation-aware behavior prediction, off-ball contribution quantification, and data-driven fatigue analysis. These capabilities have the potential to benefit multiple stakeholders: professional coaches and analysts can leverage \metrixDoubleRelativeshort~and \metrixDefcoordinationshort~to systematically identify defensive vulnerabilities and optimize stamina management strategies; and athletes may benefit indirectly through more objective, data-driven performance feedback that complements traditional coaching intuition.
However, responsible use of the dataset requires awareness of two key contextual limitations.
\paragraph{Scope of Diagnostic Metrics.} The coordination metrics, \metrixDoubleRelativeshort~and \metrixDefcoordinationshort, and the expert-annotated loss reasons are designed as diagnostic instruments for causal tactical evaluation . These metrics quantify spatial configurations and defensive burden distribution , but cannot capture unobservable factors such as psychological pressure, deliberate tactical sacrifices, or real-time decision uncertainty inherent in competitive matches. Practitioners are therefore cautioned against applying these metrics as absolute scoring rubrics in automated refereeing systems, player performance penalization, or contract evaluation contexts. We recommend that any deployment of these metrics beyond research settings be accompanied by domain expert oversight and contextual interpretation.
\paragraph{Applicability to Non-Elite Populations.} \DatasetName~is curated entirely from elite BWF World Tour tournaments, meaning the tactical dynamics encoded in the dataset reflect peak professional athletic capabilities. Without further validation on non-elite populations, directly applying these elite-derived models to amateur, youth, or collegiate-level players may not yield reliable tactical conclusions, as the underlying physical and tactical assumptions may not transfer across different competition levels. Empirical validation on non-elite populations is recommended before drawing cross-level tactical conclusions.


\end{document}